\documentclass[preprint,journal]{vgtc}            

\onlineid{1197}

\preprinttext{To appear in IEEE Transactions on Visualization and Computer Graphics.}

\ieeedoi{xx.xxxx/TVCG.201x.xxxxxxx}

\vgtccategory{Research}

\vgtcpapertype{Analytics \& Decisions}

\title{\name{}: Visualizing and Understanding Commonsense Reasoning Capabilities of Natural Language Models}

\author{%
  Xingbo Wang,
  Renfei Huang, 
  Zhihua Jin,
  Tianqing Fang,
  and Huamin Qu
}

\authorfooter{
  \item
  	Xingbo Wang is with Weill Cornell Medical College, Cornell University. This work was done at the Hong Kong University of Science and Technology.
  	E-mail: xingbo.wang@\{connect.ust.hk, med.cornell.edu\}.
  \item
  	Renfei Huang, Zhihua Jin, Tianqing Fang, and Huamin Qu are with the Hong Kong University of Science and Technology. E-mail: \{rhuangan, zjinak, tfangaa, huamin\}@ust.hk.

}

\abstract{%
Recently, large pretrained language models have achieved compelling performance on commonsense benchmarks. 
Nevertheless, it is unclear what commonsense knowledge the models learn and whether they solely exploit spurious patterns.
Feature attributions are popular explainability techniques that identify
important input concepts for model outputs. However, commonsense knowledge tends to be implicit and rarely explicitly presented in inputs. These methods cannot infer models' implicit reasoning over mentioned concepts.
We present \name{}, a visual explanatory system that utilizes external commonsense knowledge bases to contextualize model behavior 
for commonsense question-answering.
Specifically, we extract relevant commonsense knowledge in inputs as references to align model behavior with human knowledge.
Our system features multi-level visualization 
and interactive model probing and editing for different concepts and their underlying relations. 
Through a user study,
we show that \name{} helps NLP experts conduct a systematic and scalable visual analysis of models' relational reasoning over concepts in different situations.
}


\teaser{
  \includegraphics[width=.98\linewidth]{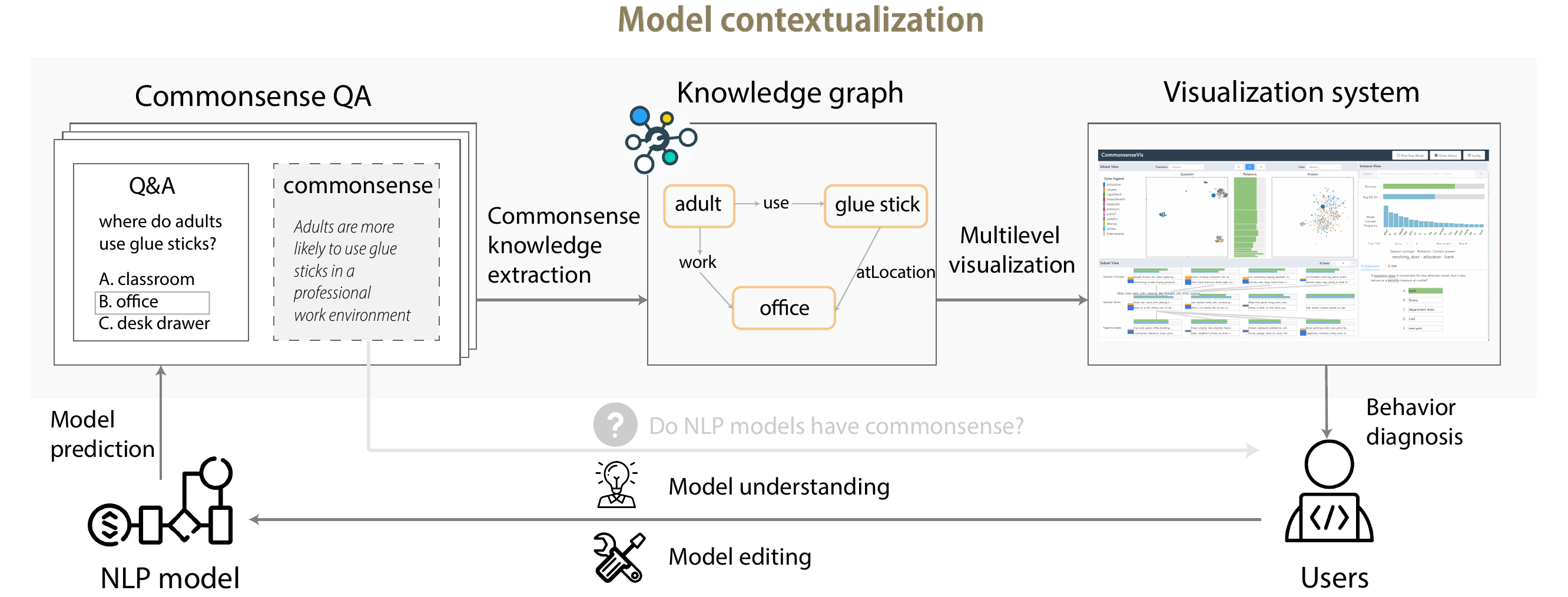}
  \vspace{-3mm}
  \caption{\vis{Since commonsense knowledge is not explicitly stated, it is challenging to conduct a scalable analysis of what commonsense knowledge NLP models do (not) learn. We employ a knowledge graph to derive implicit commonsense in the model input as context information. Then, we use it to align model behavior with human reasoning through multi-level interactive visualizations. Thereafter, users can understand, diagnose, and edit specific knowledge areas where models do not perform well.}}
  \label{fig:workflow_teaser}
}

\graphicspath{{figs/}{figures/}{pictures/}{images/}{./}} 

\usepackage{tabu}                      
\usepackage{booktabs}                  
\usepackage{lipsum}                    
\usepackage{mwe}                       

\usepackage{mathptmx}                  
\usepackage{paralist}
\usepackage{comment}
\usepackage{amsmath}
\usepackage{amssymb}
\usepackage{amsfonts}

\newcommand{\xingbo}[1]{{\color{black} #1}}
\newcommand{\renfei}[1]{{\color{black} #1}}
\newcommand{\renfeivis}[1]{{\color{black} #1}}
\newcommand{\rev}[1]{{\color{black} #1}}
\newcommand{\vis}[1]{{\color{black} #1}}
\newcommand{\todo}[1]{{\color{black} #1}}
\newcommand{\referappendix}[1]{{\color{black} #1}}
\newcommand{\revv}[1]{{\color{black} #1}}
\newcommand{\revvv}[1]{{\color{black} #1}}

\newcommand{\ie}{i.e.}
\newcommand{\eg}{e.g.}
\newcommand{\esp}{esp.}

\newcommand{\hl}[1]{\emph{``{#1}''}}
\newcommand{\imp}[1]{\textbf{\textit{{#1}}}}
\newcommand{\systemname}{{CommonsenseVIS}}
\newcommand{\name}{{\textit{CommonsenseVIS}}}
\newcommand{\gv}{Global View}
\newcommand{\sv}{Subset View}
\newcommand{\iv}{Instance View}
\newcommand{\cpn}{ConceptNet}

\begin{document}

\firstsection{Introduction}
\maketitle
\label{sec.intro}
Commonsense knowledge describes the general facts and beliefs about the world that are obvious and intuitive to most humans. It allows people to smoothly explore and reason over everyday events and situations.
For example, \hl{my parents are older than me} and \hl{take an umbrella when it rains}.
Equipping machines with humanlike commonsense reasoning abilities can benefit the development of social robots and intelligent personal agents to assist humans in daily tasks.

Commonsense knowledge and reasoning have been important and long-standing challenging topics in the natural language processing (NLP) community.
Many researchers have devoted their efforts to building commonsense knowledge bases by extracting information from existing data sources (\eg, Wikipedia) or acquiring it from domain experts or crowd workers.
Generally, the commonsense knowledge is represented as graphs, where nodes denote the conceptual entities (\eg, cars, people) and links describe the relations between different concepts (\eg, people \hl{is capable of} driving cars).
Building upon the knowledge bases, a few benchmark datasets have been designed to evaluate and improve NLP models for automated commonsense reasoning. Particularly, question answering (QA) is the primary and popular form of benchmarks~\cite{CSQA1, CSQA2, zellers2019hellaswag, bisk2020piqa, bhagavatula2019abductive, singh2021com2sense, huang2019cosmos}.

Recent advances in large pre-trained language models (PLMs) of NLP (\eg, BERT \cite{tenney2019bert}, GPT \cite{gpt3}, and T5 \cite{t5model}) have yielded impressive and even human-level performance \cite{humanparity} on commonsense benchmarks.
However, these models lack interpretability and transparency, 
which hinders model debugging and development for real-world applications. 
It is unclear what commonsense knowledge the models have learned and used in the process of reasoning,
and whether they merely explore the spurious correlation in the datasets.
This issue has led to a rallying call for explaining NLP models to reflect their real commonsense reasoning capabilities and to build more robust benchmarks and models.

To help NLP experts understand the NLP model's reasoning process, feature attribution methods (\eg, LIME~\cite{ribeiro2016should}, SHAP~\cite{lundberg2017unified}) are popular explainability techniques, which quantify the importance of input features (\eg, words and phrases) to the model outputs. 
Therefore, NLP experts can identify critical concepts for model predictions and determine whether they are aligned with human knowledge.
We have witnessed the success of these methods for various applications (\eg, sentiment analysis~\cite{chen2020generating} and fake news detection~\cite{ayoub2021combat}). 

However, feature attribution methods cannot be directly applied to explain models for commonsense reasoning tasks. 
First, \revv{\textit{they are incapable of 
revealing models' relational reasoning over concepts (\eg, entities) in different contexts}} since their relations may require background knowledge and not be explicitly presented in the input.
For example, in \hl{take an umbrella when it rains}, the inherent commonsense is that the umbrella \hl{is used for} \revv{protection from the rain}, which is not mentioned in the original statement.
\revv{Moreover, contexts significantly influence the reasoning over these implicit relations between concepts.}
For instance, depending on the weather, the umbrella can \hl{be used for} \revv{protection from the rain or sun.}
Furthermore, feature attributions often focus on individual instances. 
\revv{Given the complexity and vastness of the commonsense knowledge space the models operate on, where concepts are intertwined with various relations and contexts, it is challenging to scale up these methods to efficiently 
build high-level abstractions of model behavior (\eg, under what contexts a relation is well learned) and generalize model understanding to large datasets.}
\vis{Furthermore, to better align the model with human knowledge and expectations, it is crucial to not only understand its reasoning but also actively inject and update the desired knowledge within the model, such as using human feedback to finetune ChatGPT~\cite{DBLP:journals/corr/abs-2203-02155}.}

\revv{Visual analytics~\cite{sharedinterest, amershi2015modeltracker, whatiftool} have been an effective approach for summarizing complex data characteristics and facilitating data-driven model understanding at scale.}
Motivated by this, we design and develop a visual analytics system, 
\name{}, which enables \revv{NLP experts (\eg, model developers)} to conduct a systematic and scalable analysis of the commonsense reasoning capabilities of NLP models (outlined in \autoref{fig:workflow_teaser}). 
Going beyond many existing visual explanation tools~\cite{tenney2020language,whatiftool,amershi2015modeltracker,sharedinterest} that focus on input-output behaviors of models, our system integrates an external knowledge base to derive implicit commonsense knowledge from the input and uses them as additional contexts to align model behavior with human reasoning through interactive visualizations.
We focus on commonsense question answering (CQA), a popular task for evaluating commonsense reasoning abilities, and showcase our system on the representative CSQA benchmark dataset~\cite{CSQA1}. 
\cpn{}~\cite{conceptnet}, a commonsense knowledge graph, is used to extract commonsense knowledge from data as concept-relation triplets for model contextualizations. 
Our system provides multi-level visualizations of model behavior by comparing important input features for model decisions with the extracted triplets from ConceptNet.
At the global level, the system adopts data transformation and projection strategies to summarize model performance on questions and relations and assesses the overall relation learning.
At the subset level, the system presents a contextual summary of the alignment between model behavior and related \cpn{} knowledge for different subsets.
And at the local level, the system provides visual explanations for individual instances and allows for model probing and editing to identify and enhance specific knowledge areas where models underperform.
\vis{Through a user study using CommonsenseQA (CSQA)~\cite{CSQA1} dataset, we show that \name{} can help NLP experts effectively understand, diagnose, and edit model knowledge
on concepts and their implicit relations in different contexts.}

The major contributions of this paper are summarized as follows:
\begin{compactitem}
    \item \vis{{\name}, a visual analytics system that supports a systematic and scalable analysis of the model’s reasoning on commonsense tasks involving a large number of concepts and their relations. Particularly, it helps align model behavior with human reasoning through model contextualization, multi-level visualizations, and interactive model probing and editing.}
    \item A user study with cases that shows the effectiveness and usability of our system in revealing, diagnosing, and editing underlying commonsense knowledge the language model does not learn.
\end{compactitem}

\section{Related Work}\label{sec.related_work}

We discuss related work in commonsense reasoning, explainable AI methods, and visualization for NLP models.

\subsection{Commonsense Reasoning}\label{subsec.cs_reasoning}

Here, we introduce the most relevant work, including large knowledge graphs, benchmark datasets, and commonsense reasoning methods.


\textbf{Large-scale knowledge graphs} act as the representation of commonsense knowledge for NLP models to access and exploit. 
The commonsense knowledge graphs (CSKGs) can be divided into two categories, which are human-annotated CSKGs (\eg, ConceptNet \cite{conceptnet}, ATOMIC \cite{atomic2019, atomic2020}, and GLUCOSE \cite{glucose}) and web content extracted CSKGs \cite{webchild, transomcs}. 
\cpn~\cite{conceptnet} is a comprehensive large-scale knowledge graph with over 3.4M entity-relation tuples to connect concepts (words and phrases) by 36 types of relations. It primarily focuses on taxonomic, lexical, and physical knowledge. 
It is collected by crowdsourcing and merged with high-quality knowledge databases.
Here, we use \cpn{} to reveal commonsense knowledge in data instances and contextualize model behavior.
\textbf{Benchmark datasets} for evaluating NLP models' commonsense reasoning abilities typically involve question-answering tasks, reading comprehension~\cite{huang2019cosmos, zellers2019hellaswag},  open-ended question answering~\cite{boratko2020protoqa, lin2020differentiable}, and multiple-choice questions~\cite{socialiqa, bisk2020piqa, sakaguchi2021winogrande, CSQA2,   singh2021com2sense}.
One example is ComonsenseQA (CSQA) dataset, which consists of 12k commonsense questions authored by crowd workers in a 5-way multiple-choice format. Among the five answer choices, three are directly extracted from ConceptNet, with one being the correct answer. Then the crowd workers create two additional distractors, one from ConceptNet and another authored by themselves. 
CSQA evaluates models mainly on factual and physical commonsense relations (\eg, \texttt{atlocation}) between entities. And we use CSQA to showcase how our system enables scalable and systematic analysis of NLP models' commonsense reasoning abilities.

\vis{\textbf{Commonsense reasoning models} include
\revv{large language models (LLMs)~\cite{t5model,gpt3,unicorn, 2020unifiedqa, khashabi2022unifiedqa} pretrained on large text corpora.
They achieve impressive performance on commonsense benchmarks. Nevertheless, these models exhibit limitations in their capacity to possess and effectively utilize commonsense knowledge for reasoning tasks~\cite{bian2023chatgpt,ma2021knowledge}.}}
\vis{To enhance models' commonsense knowledge, some methods~\cite{lin2019kagnet,yasunaga2021qagnn, humanparity}
integrate external knowledge bases and/or linguistic theories into the models to provide more contexts and facts for improving model accuracy.}
\vis{In addition, \revv{pretrained} language models can be used as knowledge bases to generate clarification questions~\cite{shwartz2020unsupervised}, commonsense explanations~\cite{rajani2019explain}, and prompts~\cite{liu2021generated} to enhance commonsense reasoning.}
\vis{However, they do not explain what commonsense knowledge is injected (un)successfully. 
\revv{And we present a model-agnostic explanation system to systematically evaluate commonsense knowledge that these NLP models possess and utilize for reasoning tasks.}
}

\subsection{Explainable AI Techniques for NLP} 
\label{subsec.xai}
Explainable AI (XAI) is critical to promote model transparency and reliability~\cite{xai_survey}. We focus on post-hoc model explanations via a model-agnostic approach. 
One popular \revv{post-hoc} explainability technique is feature attribution~\cite{ribeiro2016should, lundberg2017unified, sundararajan2017axiomatic}, which quantifies the contribution of input features to the model output. 
We use a model-agnostic method, SHAP~\cite{lundberg2017unified}.
Another related direction is counterfactual analysis~\cite{kaushik2019learning, wu2021polyjuice}, which uses examples to reverse the target label, helping understand model decision boundaries.
Our system enables question manipulation to probe model behavior regarding specific concepts or relations.

While previous work~\cite{li-etal-2022-systematic, zhou2020evaluating} has explained Natural Language Processing (NLP) models via zero-shot or few-shot accuracy evaluations on pre-trained language models, our research conducts a detailed, systematic analysis of model behavior on diverse commonsense concepts and relations. 
Compared to other methods~\cite{petroni2019language, tenney2019bert, reif2019visualizing, cui2020commonsense} that design auxiliary classification tasks to understand linguistic knowledge in NLP models, we aim to reveal the role of commonsense knowledge in the model's reasoning process. 
\revv{Directly prompting large language models helps probe the simple facts embedded in them. However, many NLP models~\cite{2020unifiedqa, unicorn, lin2019kagnet} cannot be easily prompted due to their inherent designs.
Finally, some studies~\cite{meng2022locating, meng2022mass, dai-etal-2022-knowledge} conduct causal analysis that attributes a piece of knowledge to specific neurons in the models. However, these methods do not efficiently summarize how a model learns different commonsense knowledge.} Our system adopts the model-agnostic feature attribution method to quantify model behavior and contextualizes it with a knowledge graph. It then employs multi-level visualizations to facilitate systematic exploration of model behavior across various commonsense concepts and relations.

\subsection{Visualization for Understanding NLP Models}
\label{subsec.vis4ai}
Visualizations can effectively help understand NLP models~\cite{adadi2018xaisurvey}.
Model-specific visualizations~\cite{strobelt2017lstmvis,ming2017understanding,vig2019bertviz,hoover2019exbert,jin2022gnnlens} reveal the model's inner workings, such as the behavior of neurons, layers, and attention maps. 
By examining these visualizations, users can gain insights into hidden state dynamics~\cite{strobelt2017lstmvis}, the relationships between hidden states and words~\cite{ming2017understanding}, and diagnose model bias~\cite{vig2019bertviz}.

\revv{Many model-agnostic visualizations}~\cite{whatiftool, sharedinterest, mediators, amershi2015modeltracker, wang2021m2lens, liang2022multiviz} focus on input-output model behavior and are generally applicable to different models.
For example, What-If Tool \cite{whatiftool} allows users to understand model behavior concerning feature importance, different inputs, and other hypothetical situations.
M\textsuperscript{2}lens~\cite{wang2021m2lens} characterizes intra- and inter-modal interactions learned by multimodal models.
Shared Interest \cite{sharedinterest} compares the reasoning of models and humans using saliency results and ground truths. 
\revv{DeepNLPVis \cite{li2022unified} and MultiViz~\cite{liang2022multiviz} present multi-level visualizations to explore both the behavior and working mechanisms of different NLP models across different tasks.} 


However, these studies do not provide insights into the commonsense knowledge that models may (not) learn. To fill the gap, we propose a model-agnostic approach that leverages multi-level visualization and an external knowledge base to contextualize
the implicit reasoning of models over concepts and relations in commonsense questions.

\section{Design Requirements}
\label{sec.design_requirements}
We aim to develop a visual analytics system to help NLP experts understand and diagnose commonsense reasoning capabilities of NLP models in a systematic and scalable manner.
Explaining such model abilities helps users determine whether models are suitable and trustworthy for downstream applications and enhance specific knowledge that models do not learn well.
However, it is challenging to depict and summarize the vast and complex space of commonsense knowledge that models learn, as it is not directly presented in the input, and concepts are entangled with various relations and contexts.

We first conducted a literature review on explainability techniques \renfei{\cite{ribeiro2016should, lundberg2017unified, kaushik2019learning, wu2021polyjuice}} and visual analytics~\renfei{\cite{sharedinterest, whatiftool, feng2023xnli,jin2023shortcutlens}} for NLP, and commonsense reasoning~\renfei{\cite{conceptnet, cui2020commonsense}}. 
To further characterize users' common practices and needs, we collaborated with three NLP experts (\imp{E1-E3}, \imp{E1} is the coauthor) through regular weekly meetings for about six months.
\imp{E1} is a Ph.D. candidate who investigates commonsense knowledge acquisition and reasoning. 
\imp{E2} has obtained a Ph.D. degree in HCI and has rich experience in building human-centered interactive NLP models.
And \imp{E3} is a research scientist from an international media company whose expertise is in explainable AI and visualization for NLP. 
During the meetings, we asked them about 
1) the general methods of NLP model evaluation; 
2) what types of explanations for models' commonsense reasoning capabilities;
and 3) the desired system task support.
Meanwhile, we developed our system prototypes iteratively and collected their feedback for further improvement.


\rev{\textbf{Current practice and limitations.}}
Our users usually start with performance metrics (\eg, accuracy) to locate data instances (\esp, wrong predictions) and manually summarize what commonsense knowledge is needed for inference. 
Specifically, users identify important relations and concepts for commonsense reasoning and combine performance metrics with feature attribution methods to determine whether models capture important concepts or superficial word associations. Moreover, they can probe the models by modifying the data instances to verify their hypotheses.
However, this analysis process is tedious, mentally demanding, and difficult to generalize to larger data subsets. They desire a visual analytics tool to analyze what commonsense knowledge is contained in data instances and (not) learned by NLP
models.

\textbf{R1. Reveal commonsense knowledge in data instances.}
Our users need to distill the external commonsense knowledge from data instances, which helps verify if model behavior aligns well with human knowledge~\cite{sharedinterest,jin2023shortcutlens}. Since concepts and relations are critical components of commonsense knowledge~\cite{conceptnet, cui2020commonsense}, the system should extract relevant concepts and their relations in questions as references to understand data itself and model behavior:

\begin{compactdesc}
\setlength\itemindent{-1em}
  \item[Q1:] What concepts (\eg, entities) are mentioned in the instances?
  \item[Q2:] What are the latent relations between the mentioned concepts?
\end{compactdesc}


\textbf{R2. Summarize model performance on different concepts and relations.}
Our users usually depend on accuracy scores to pinpoint cases where models fail and prioritize exploring them.
To scale up the analysis of individual instances to large datasets, it is necessary to summarize model performance from multiple aspects~\cite{whatiftool, sharedinterest, feng2023xnli}.

\imp{E3} said that a concept-driven summary can reveal what topics models perform well. \imp{E1} mentioned that compared to the vast concept space, relations have more summative power and connect concepts meaningfully. \imp{E3} suggested relating model performance to linguistic contexts to assess their ability to use commonsense knowledge in different situations. For instance, testing models on instances where adults and children use staplers helps understand whether models can distinguish between them.
Therefore, the system should answer:


\begin{compactdesc}
\setlength\itemindent{-1em}
  \item[Q3:] What concepts, relations, and their combinations are predicted right or wrong by the models?
  \item[Q4:] What are the contexts of the relations and concepts? What is the model performance?
\end{compactdesc}



\textbf{R3. Infer model relational reasoning over concepts based on relevant commonsense knowledge.}
To develop a mental model about models' commonsense knowledge and reasoning, users need to first use their own prior knowledge to build the relevant reasoning paths that connect important words in statements. Then, they need to check whether models capture these meaningful concepts in statements based on their importance to model predictions. Although sometimes models are correct, they may rely on task-unrelated linguistic features (\eg, stop words) to make decisions. 
Moreover, \imp{E2} and \imp{E3} thought that to better surface the patterns of how models regard unmentioned relations, it is necessary to show whether models attach importance to the mentioned words in statements connected by those relations.
By concept-driven comparison between important concepts recognized by models and humans, users can generate hypotheses about models' relational reasoning over concepts:

\begin{compactdesc}
\setlength\itemindent{-1em}
  \item[Q5:] What concepts are important for model predictions? Are they reasonable?
  \item[Q6:] What unexpressed relations are necessary for inference? Do models cover the concepts connected by these relations?
  \item[Q7:] What are the differences between the important concepts for commonsense reasoning and for model predictions?
\end{compactdesc}




\textbf{R4. Allow interactive probing and editing of NLP models.}
One straightforward and useful way to understand and debug models is interactively interrogating them~\cite{whatiftool, kaushik2019learning, wu2021polyjuice}. To generate and verify the what-if hypothesis about model behavior, users can conduct counterfactual analysis by manipulating specific input components and seeing how models react to these changes.
This helps disentangle influences of individual concepts in statements for model predictions and check whether models are biased towards some concepts. Moreover, modifying the input components can test the robustness of models against noisy concepts and probe the underlying relations of interest that link the mentioned concepts in the input.
\vis{After model probing, users may desire to conduct posthoc editing of model behavior to inject their desired knowledge and make a flexible localized update about specific knowledge areas that models do not learn well~\cite{de2021editing,mitchell2022fast}.}

\revvv{Given the requirements, we consolidated a series of system tasks that guides the systematic exploration of models' commonsense reasoning capabilities: Initially, commonsense knowledge from data is extracted as concept-relation triplets (\imp{R1}). 
Then, the system summarizes model performance across these concepts and relations
(\imp{R2}), and further assesses the overall relation learning (\imp{R3}). 
Next, it enables users to pinpoint error instances related to specific concepts or relations (\imp{R2}). 
For these instances, the system summarizes the concepts considered important by models in varying contexts, aligning them with those in the extracted triplets (\imp{R3}). 
Moreover, visualization is utilized in conjunction with model probing (\imp{R4}) to comprehensively explain models' input-output behavior on these instances (\imp{R3}).
Finally, users can bookmark instances for targeted model refinement (\imp{R4}).
}

\section{System \& Methods}
\label{sec.system}

\vis{
Our system, \name{}, leverages an external knowledge graph to summarize and derive commonsense knowledge and facilitate multi-level exploration and diagnosis of model behaviors in commonsense question-answering tasks. We focus on question-answering tasks because they are a common evaluation method for natural language understanding, and most commonsense reasoning benchmarks adopt the QA format~\cite{socialiqa, bisk2020piqa, sakaguchi2021winogrande, 2020unifiedqa, CSQA1, CSQA2}.}

\begin{figure*}[t]
\centering
    \includegraphics[width=.95\textwidth]{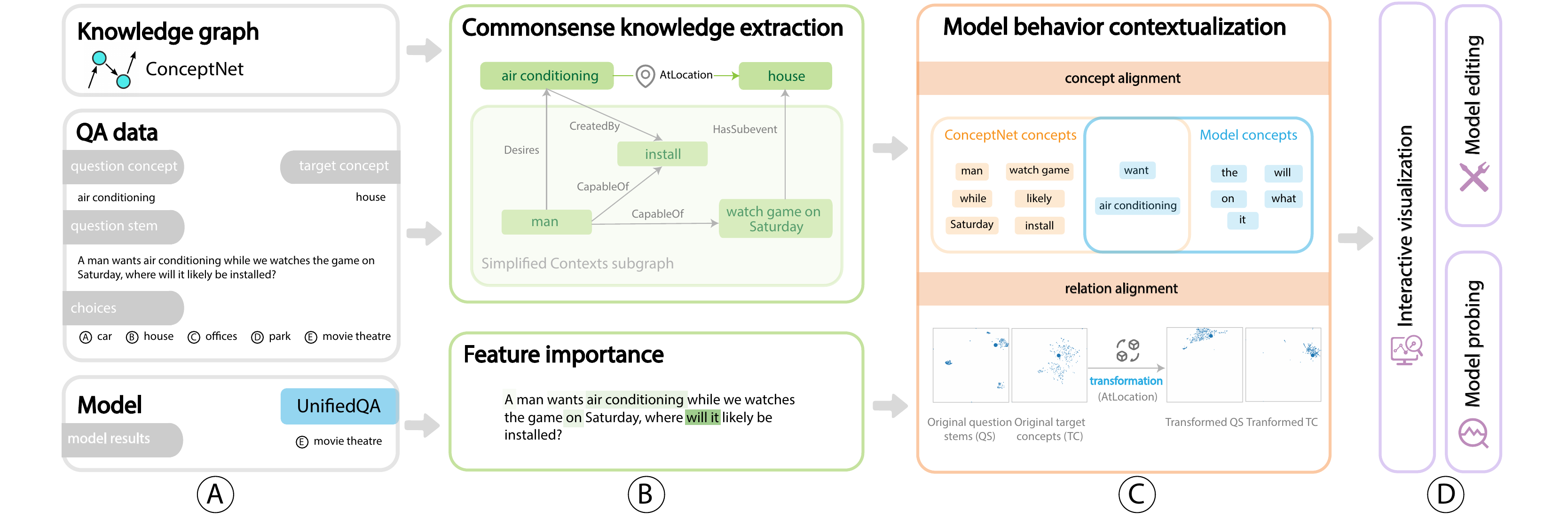}
    \vspace{-4mm}
    \caption{\renfei{\revv{The system overview.} (A) First, the QA data, NLP model, and knowledge graph are input into the system. (B) Then our system extracts commonsense knowledge in data input based on \cpn{}, and calculates the feature importance scores for individual words in questions. (C) We then align model behavior with ConceptNet knowledge regarding various concepts and underlying relations. (D) Finally, the results are integrated into the interactive visualization system with extended support of model probing and model editing.}}
    \label{fig:system_framework}
    \vspace{-4mm}
\end{figure*} 

\revv{\subsection{System Overview}}
\label{subsec.system_framework}
\revv{\autoref{fig:system_framework}} provides an overview of our system. \name{} takes in QA instances and an NLP model to compute model answers. Then, it identifies important concepts (\ie, words) in questions using feature attribution methods and extracts relevant commonsense knowledge from input data using an external knowledge base. This knowledge helps align the model behavior with \cpn{}. The user interface enables multi-level exploration, interactive probing, and editing.

\subsection{System Data \& Model}\label{subsec.system_data}
\rev{Here, we introduce the system input, including the QA data, model, and external knowledge base for contextualizing model behavior.

\textbf{QA data.} Each QA instance contains a \emph{question concept}, a \emph{target concept (\ie, answer)}, \emph{alternative answers} (if any), and a \emph{question stem}.
Following the previous commonsense QA benchmarks~\cite{CSQA1,CSQA2,atomic2019}, \revv{\textbf{concepts} are defined as words, and question stems provides \textbf{contexts} for the commonsense relations between the question and target concepts.}
\revv{As shown in \autoref{fig:system_framework}A, the question concept is air conditioning, the target concept is house, and air conditioning is located at the house. And the context in the question stem is: ``A man...watches the game on Saturday...''.}
If a question concept is absent,
knowledge graph embedding methods~\cite{bordes2013translating} can be used to determine the relation strength between the target concept and words in the question stem. The word with the highest score becomes the question concept~\cite{lin2019kagnet}.

We utilize one representative commonsense QA benchmark, CSQA~\cite{CSQA1}, for demonstration.
The dataset has 12,102 multiple-choice questions, covering diverse topics and various forms of commonsense.
Each human-authored question contextualizes relations between a question concept and a target concept (\ie, the correct answer among five candidates). 
Triplets of these concepts and relations are drawn from \cpn{}~\cite{conceptnet}.
The most frequent question concepts are about people, water, and animals, probing various relations such as spatial (41\%) and causal (23\%).
Questions are formulated in diverse forms (\eg, wh-questions, statements, and hypotheses) with 13 words on average.

\textbf{QA model.}
Our system is designed to accommodate various NLP models that select answers to given questions, as it focuses on the input-output model behavior and we can adopt model-agnostic feature attribution methods to quantify this behavior.

\revv{For the purpose of system demonstration, we have chosen UnifiedQA\footnote{\small \url{https://huggingface.co/allenai/unifiedqa-v2-t5-large-1363200}}~\cite{2020unifiedqa, khashabi2022unifiedqa} as an example for model analysis. It is an open-source, general QA model that has been pre-trained across various QA datasets, showing great generalization capabilities.
We use SHAP to compute the importance scores of model inputs because of its strong theoretical foundation and widespread adoption in various domains.}



\textbf{Commonsense knowledge base.}
We utilize an external knowledge base to capture the commonsense knowledge in the QA data, which provides context for inferring the model's implicit reasoning.  To ensure meaningful and helpful context for model analysis, the knowledge base must \textit{sufficiently cover relevant commonsense }reflected in the QA data. 

We adopt \cpn{}~\cite{conceptnet} as an external resource, a large-scale commonsense knowledge graph connecting \textbf{concepts} (\ie, words) with \textbf{relations}.
The graph integrates diverse knowledge sources with over 8 million nodes and over 21 million links.
Particularly, it uses 36 general relations (\eg, \hl{IsA}, \hl{UsedFor}) to connect words, mostly covering taxonomic, lexical knowledge, and physical commonsense knowledge.
ConceptNet is widely used to enhance NLP models with commonsense capabilities~\cite{lin2019kagnet, feng2020scalable} and build reasoning benchmarks~\cite{CSQA1, CSQA2, lin2021riddlesense}. 
For example, \revv{the questions and answers in CSQA are based on word-relation triplets (\texttt{A}, \texttt{Relation}, \texttt{B}) from ConceptNet.
The prevalent relations include \texttt{AtLocation} (A is typically located at B), \texttt{Causes} (A is the typical cause for B), and \texttt{CapableOf} (A can typically do B).}
Moreover, over 98\% of words in CSQA questions are covered in ConceptNet. Therefore, ConceptNet is a suitable resource for contextualizing model behaviors on CSQA and other commonsense QA datasets~\cite{feng2020scalable}.

}

\subsection{Extract Relevant Commonsense Knowledge}\label{subsec.extract_commonsense}
To help users build a concrete understanding of commonsense questions and their connections with model behavior, we distill relevant commonsense knowledge in data instances based on \cpn{} (\imp{R1}).

The commonsense knowledge extraction consists of two major steps \renfei{(\autoref{fig:system_framework}B)}, including recognizing mentioned concepts in the questions and constructing sub-graphs on the concepts.
\rev{To reflect the reasoning paths from the question concept to the target concept/answer, we perform tokenization of the question stem by n-gram (\rev{$n=1,2,3$\footnote{\rev{To balance the coverage of meaningful phrases with varying lengths and computational complexity, we limit maximum gram size to be three~\cite{manning1999foundations}.}}}) and match the tokens (\ie, words of length $n$) with the concepts in ConceptNet to identify a set of candidate concepts for commonsense reasoning. 
Since the matched concepts (with different lengths) may have overlaps, we reduce the redundancy by keeping the longest matched concepts in \cpn{}.}
Moreover, to enhance the robustness of matching, we conduct soft matching by lemmatization and removal of stop words and punctuations.
For example, after token matching, the candidate concepts in a question \hl{A man wants air conditioning, ...} will be \{man, want, air conditioning, ...\}.
Next, those tokens are used to construct a knowledge graph that contains the question concept and the target concept
to describe the reasoning process.
By leveraging the connections among the candidate concepts, question concept, and target concept in \cpn{}, we establish relational paths, employing a two-hop relation search. 
We set the hop size to two to balance the computation scalability and coverage of reasoning paths, following the prior work~\cite{lin2019kagnet,yasunaga2021qagnn}.
Thereafter, the resulting graph of concepts and relations (in~\renfei{\autoref{fig:system_framework}B}) describes the relevant commonsense knowledge for the question. This graph is referred as ConceptNet knowledge.

\subsection{Align Model Behavior with ConceptNet Knowledge}\label{subsec.align_model}
\rev{To help users build mental models about the model's relational reasoning over concepts,
we align the model input-output behavior with ConceptNet knowledge regarding different concepts and relations (\imp{R3}).}
For concept alignment \renfei{(\autoref{fig:system_framework}C)}, SHAP is used to calculate the importance scores of the input concepts to the model outputs.
And we call those with large positive influences on the model predictions as model concepts.
\rev{Then, we compute the differences between the set of model concepts and the set of \cpn{} concepts (\ie, question concepts and concepts in question stems derived in \autoref{subsec.extract_commonsense}).}
\rev{For relation alignment, 
we mainly consider the key relations (\ie, the relations between question concepts and target concepts) for correctly answering the questions (\autoref{fig:system_framework}C). Noticing that \textit{question concepts} are included in question stems as \textit{model inputs} and \textit{target concepts are ground truths for model outputs}, we surface the model learning of 
their relations
by investigating the relationships of model inputs and outputs.}
\rev{Specifically, the inputs and outputs are high-dimensional embeddings that the model operates on. And we compute the linear transformation matrix $W \in \mathbb{R}^{d \times d'}$ between model input embeddings $X \in \mathbb{R}^{N \times d}$ and output embeddings $Y \in \mathbb{R}^{N \times d'}$.
Particularly, to reflect relations between question concepts and target concepts encoded in $W$, we use those correctly-predicted instances (\ie, model predictions $P$ are equal to target concepts) as the anchor points for the transformation. And we adopt a least-square error objective to compute the linear matrix $W$: $\mathop{\mathrm{argmin}}_{W \in \mathbb{R}^{d \times d'}}~||XW - Y||_2$, where $(X, Y) = \{(x_i, y_i)~|~TC_i = P_i\},~i=1,...,N $.
}

\rev{The general idea is that the input-output relationships can be modeled by translations in the model embedding space~\cite{bordes2013translating, dinu2014improving}: if a model can capture the relations between question concepts and target concepts,  then question concept embeddings transformed with the matrix $W$ should be close to target concept embeddings.}

\subsection{Model Editing}\label{subsec.model_editing}

\vis{After identifying model deficits in specific commonsense knowledge, we present \textit{editor networks} to modify model parameters that can correct problematic model answers (\imp{``reliability''}), as well as other semantically-equivalent questions (\imp{``generality''}) without affecting unrelated knowledge much (\imp{``locality''}).
Particularly, editor networks are neural networks trained to modify model parameters (from $\theta$ to $\theta'$) with the objectives that maximize editing accuracy on both editing targets ($x_e, y_e$) and their equivalence ($x_e', y_e'$) while minimizing differences (KL divergence) in model predictions on locality examples ($x_{loc}, y_{loc}$) before and after the edits: $L_e = -log p_{\theta}'(y_{e}'|x_{e}'), L_{loc} = \texttt{KL}(p_{\theta}(\cdot | x_{loc})||p_{\theta}'(\cdot | x_{loc}))$.
The total loss is $L_{total} = - w_{e} \cdot L_{e} + L_{loc}$, where $w_e$ is a weight factor. The editing examples come from QA pairs in CSQA train/val set, 
where their equivalences are generated by popular 
data augmentation techniques, i.e., back-translation and EDA~\cite{DBLP:conf/emnlp/WeiZ19}.
Locality examples are independently sampled. We adopt gradient decomposition techniques~\cite{mitchell2022fast} to train editor networks on the last two transformer layers of the model. More technical details are included in 
\referappendix{Suppl. A.}
}

\subsection{User Interface of \systemname}
\label{sec.user_interface}

\begin{figure*}[ht]
\centering
    \vspace{-5mm}
    \includegraphics[width=\linewidth]{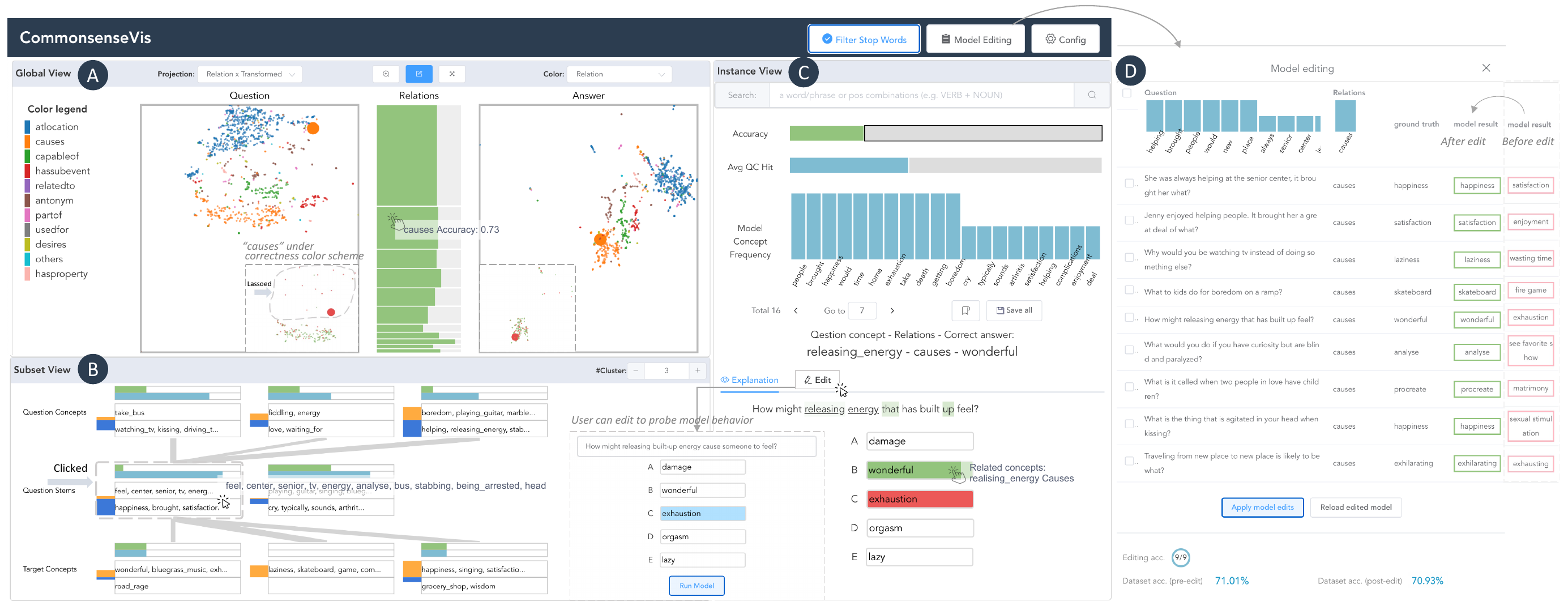}
    \vspace{-5mm}
    \caption{\renfei{The user interface of \name{}: The \gv{} (A) summarizes the model performance by the projection plots of question stems and target concepts and the relations between them. The \sv{} (B) summarizes the context alignment between model behavior and \cpn{} knowledge over different subsets. The \iv{} (C) provides the statistics and detailed local explanations of instances selected using \gv{} or \sv{}. 
    The current instance is highlighted as larger points in the \gv{}. The \iv{} also enables users to probe the model by editing the questions directly. Furthermore, users can bookmark instances and edit the model in the Model Editing Panel (D).}}
    \label{fig:teaser}
    \vspace{-4mm}
\end{figure*}

The user interface (\vis{\autoref{fig:teaser}}) enables a multi-level exploration of model behavior following an \textit{overview-to-detail} flow, contextualized by \cpn{}. 
\revv{The exploration process starts with the \gv{}, which summarizes model performance on different concepts and relations and assesses overall relation learning. 
Users then can pinpoint error cases, and the system summarizes the contexts of alignment between model behavior and \cpn{} on different subsets.
Upon selecting instance subsets in the \gv{} or \sv{}, \iv{} shows statistics and visual explanations for these instances. It facilitates interactive model probing for comprehensive understanding and enables users to bookmark particular instances for targeted model refinement.}
The system uses red to encode the model error, green to indicate accuracy, categorical colors to encode different relations and statistics.

\begin{figure}[ht]
    \centering
    \includegraphics[width=\linewidth]{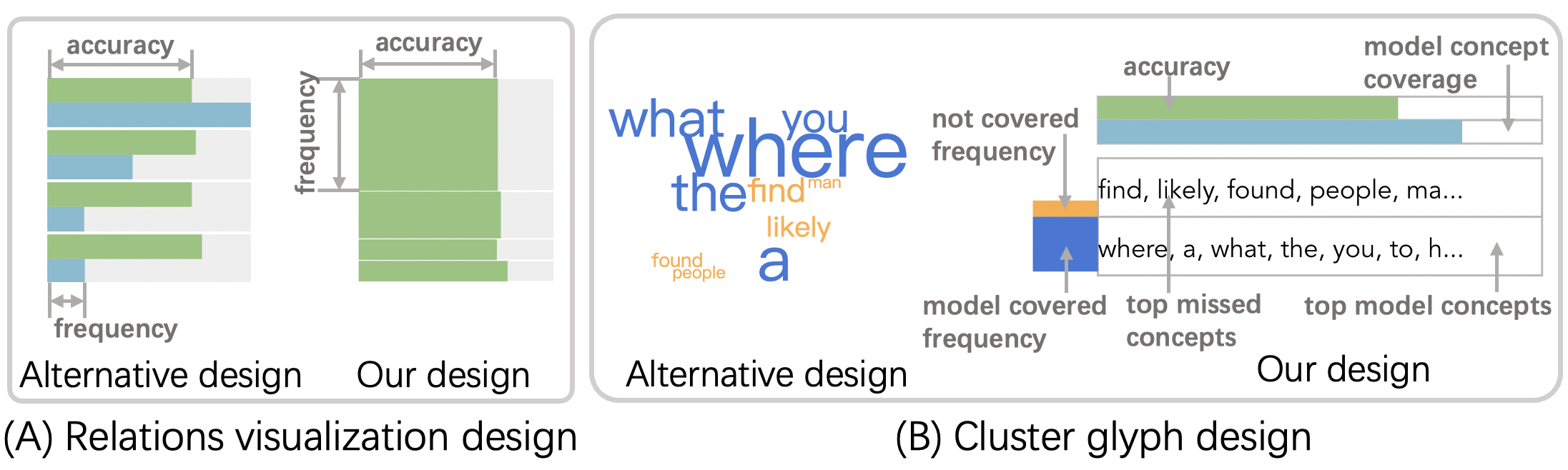}
    \vspace{-3mm}
    \caption{Alternative designs in the system for QC-TC relations visualization (A) and  the cluster glyph in the \sv{} (B). 
    }
    \vspace{-6mm}
    \label{fig:alternative designs}
    
\end{figure}

\subsubsection{\gv{}}\label{subsec.global_view}
\rev{Initially, users can refer to the \gv{} to gain an overview of the model performance regarding different concepts and commonsense relations contained in QA data (\imp{R1, R2}).}
Specifically, the \gv{} (\vis{\autoref{fig:teaser}A}) adopts different projection strategies to group question stems and target concepts (\ie, answers) and visualize them as two separate scatter plots. 
\revv{For projection, we choose UMAP~\cite{mcinnes2018umap} with cosine similarity measures to cluster model embeddings for question stems and target concepts}
because of its good processing speed and preservation of embeddings' global structure. 
\revv{After projection, similar question stems (\ie, similar contextualizations of question concepts) or target concepts are close to each other.}
\revv{To further analyze error and relation distributions among these instances, users can adjust the dot color schemes at the header. When the ``Correctness'' scheme is selected, dots are colored in red and green to show distributions of incorrect and correct instances. Alternatively, selecting the ``Relation'' scheme applies categorical colors to the dots, highlighting instances with different relations.}
Meanwhile, users can change projection mode into ``Correctness'' or ``Relation'' at the header to accentuate the differences between instances with high and low errors or instances with varied relations.
To achieve this, we utilize instance correctness and relations between question concepts and target concepts as additional labels for UMAP to perform supervised dimension reduction for clear cluster separation in the scatter plots.
\revv{To mitigate the overplotting in the scatter plots, the system supports semantic zooming that allows users to navigate specific areas of interest (\eg, error instances) within dense data points.
Moreover, users can filter out the instances with particular relations by clicking the rectangles between the two scatter plots, where each rectangle encodes relation frequency and accuracy.}
\revv{For each rectangle, we use green (instead of categorical colors) to emphasize model accuracy for that relation, where the width of the green bar denotes accuracy, and its height corresponds to relation frequency.}
The system sorts these rectangles by relation frequency, allowing users to prioritize model performance exploration of more prevalent relations.

\rev{Besides, the \gv{} assesses how the model regards latent relations between questions and answers (\imp{R3}).} In the ``Relation X Transformed'' projection mode (in \autoref{fig:teaser}), the \gv{} separates instances with different relations in the scatter plots and supports the alignment and comparison of transformed question stems with target concepts.
If there is a good correspondence between transformed clusters of question stems and target concepts in the scatter plots, then the relations between question and target concepts could possibly be learned.
\rev{Finally, users may lasso a group of instances or click specific relation bars to inspect the context summary in the \sv{}.}


\textbf{Alternative design}. We have considered an alternative---grouped bar charts---to visualize the relations between question and target concepts (\renfei{\autoref{fig:alternative designs}A}). 
For each relation, green bars show accuracy while blue bars encode frequency. The longer the bars, the larger the encoded values.
We collected experts' feedback on this alternative.
\imp{E1} said that our final design using a single color looks simpler and cleaner. 
\rev{\imp{E3} commented that horizontally aligning green bars next to blue bars in the grouped bar charts could be confusing since the frequency and accuracy have different units of measurement.
\imp{E2} reported that our final design can reflect the proportion of different relations in the whole dataset more clearly.
In addition, it sorts the frequent relations in descending order, helping prioritize the exploration.}


\subsubsection{\sv{}}\label{subsec.subset_view}
\revv{After selecting a group of instances with specific concepts or relations in the \gv{},}
users can utilize the \sv{} to explore the
concept alignment between the model behavior and \cpn{} knowledge across different subsets (\imp{R2}).
This view employs cluster glyphs to analyze model behavior across instances with semantically similar question concepts, question stems, and target concepts. Hierarchical clustering of ConceptNet Numberbatch embeddings~\cite{conceptnet} is performed for question stems, question concepts, and target concepts.
We use ConceptNet Numberbatch embeddings because they encode word meanings based on \cpn{}'s semantic network and perform well on \revv{word-relatedness benchmarks~\cite{conceptnet}.} Then, users can scan through the cluster glyphs 
and sneak peek into the corresponding model performance, the important words for model decisions, and how they are aligned with \cpn{} concepts (\renfei{\autoref{fig:alternative designs}B}).
For each cluster glyph, two bars are presented at the top \revv{showing the average accuracy (between 0 and 1) and overlap ratio (between 0 and 1)} between model concepts and \cpn{} concepts.
The lower parts display the differences between the model and \cpn{} concepts.
The first row displays the top ConceptNet concepts frequently missed by the model. And an orange bar is put to the left, revealing the frequency. Then, the second row shows the frequent model concepts and their frequency (with blue bars). 
To further explore concept associations across different questions, and question and target concepts, their cluster glyphs are connected with links if their data instances overlap---the wider the link, the greater the shared data instances.
\revv{To reduce the visual clutter of links, the system allows users to adjust the cluster numbers at the header. When users hover over a specific cluster glyph, the system highlights only the connections relevant to that cluster, while keeping other links hidden.}

\textbf{Alternative design}. 
Initially, we considered using a word cloud (\renfei{\autoref{fig:alternative designs}B}) to summarize the most frequent concepts (not) covered by the model. And the word size relates to frequency. 
However, the word cloud is not space-efficient and mixes the model concepts with ConceptNet concepts and increases the visual complexity, making the system less user-friendly.
More importantly, our users prioritize reading the concept words in plain style. Therefore, we chose our current design.


\subsubsection{\iv{}}
After selecting instances in the \gv{} or \sv{}, the Instance View \renfei{(\autoref{fig:teaser}C)} provides statistics and local explanations about the model. It enables probing of the model with different inputs and outputs to test its learning of concepts and commonsense relations (\imp{R1, R4}). 
The top stacked bars show model accuracy and average question concept (QC) hit ratio \revv{(between 0 and 1)}. Users can click the green (or gray) segments with the stacked bars to filter the data instances correctly (or wrongly) predicted. 
The histogram below displays the top frequent concepts considered important to the model. 
Users can explore individual instances and model explanations with pagination. The question stems that strongly contribute to model outputs are highlighted with green backgrounds, and question concept is underlined. Model choices colored red indicate a wrong answer, and the ground truth is colored green. 
Users can verify and generalize their findings by searching for linguistic patterns in data instances that contain certain words or structures (\eg, \hl{many NOUN}) at the top. 
\revv{For instance, after searching a question concept of interest, users can review the model performance on different contextualizations (\ie, question stems) of that concept and associated relations in the \gv{}. Then, further detail can be investigated, including statistics and model explanations for individual instances, in the \iv{}.}

\rev{For individual instances, users can edit them to form and validate hypotheses about the model learning of relations.
For example, 
if the model is wrong, users may hover over different answer choices to see their relations with \cpn{} concepts in question stems.
If both model answers and target concepts share the same relations with question concepts, the model potentially does not understand the contexts.
Then, users can edit the text content of question stems and individual answer choices (\eg, remove some words in questions and change answer choices), followed by re-running the model on the edited QA pairs. The new model answers will be highlighted in blue. By examining the new results, users can validate whether the relations between the question and target concepts are learned. 
}

\vis{Users can bookmark instances about specific knowledge that the model does not learn well. Then, they can conduct model editing in the Model Editing Panel \todo{(\autoref{fig:teaser}D)}, where information about questions, relations, ground truths, and model results are summarized in a table. Users can apply editing to instances of interest and inspect the editing performance. Moreover, they can load the edited model for exploration.}

\section{Evaluation}
\label{sec.evaluation}

We conducted a user study to evaluate how {\name{}} helps experts analyze the commonsense reasoning abilities of language models.
Specifically, we invite \xingbo{10} experts (\imp{E4-E13}) to evaluate our system on a commonsense reasoning benchmark.
The experts are NLP researchers or practitioners, and all of them have rich experience in natural language understanding topics.
We delay the introduction of their backgrounds in \autoref{subsec.user_study}.
We evaluate the system by using a state-of-the-art QA model UnifiedQA~\cite{2020unifiedqa} 
and the CSQA~\cite{CSQA1} validation set.
The CSQA validation set contains 1,221 multiple-choice commonsense QA instances, and the model performance is \xingbo{71.00\%}. \revv{We also randomly sampled 100 instances from the validation set to evaluate the commonsense coverage of \cpn{} for CSQA. For each instance, an NLP expert (\imp{E14}, not a co-author) from a tech company assessed if the relational paths extracted from \cpn{} accurately covered the necessary commonsense knowledge to answer the questions. 
The results show that ConceptNet knowledge covered the necessary commonsense in 91 out of 100 instances. More details are in Suppl. C.}

\vis{Next, we present cases of using \name{} for model analysis. The cases were found by \imp{E4} and \imp{E5} during their system exploration of model behavior in the user study. Afterward, we summarize the user behaviors under the characterized system workflow. 
And we report users' ratings and feedback on the system designs and workflow.}

\rev{\subsection{Cases of Using \systemname{}}}
\label{subsec.case_studies}

\vis{
Using \name{}, experts discovered that the model has learned the relation \texttt{atlocation} in the context of ``office'' and ``room'' properly (details are in \referappendix{
Suppl. D.1).}
However, it has limitations in \texttt{cause} relation learning and ``movie'' context understanding.}
\revvv{\subsubsection{Reveal Model Deficiencies in \texttt{Cause} Relation Learning via Multi-level Exploration and Instance Editing}}
\label{subsec.case2}

\textbf{Global Summary}
(\imp{R1, R2}) 
\renfeivis{E4 first diagnosed the model learning of \texttt{causes} relation by clicking the second largest bar between the scatter plots (\renfei{\autoref{fig:teaser}A}).
}
He selected \textit{``Relation X Transformed''} projection scheme with the \textit{``correctness''} coloring to examine the alignment between question stem and target concept clusters considering model performance (dashed area in \renfei{\autoref{fig:teaser}A}).
In the left scatter, \imp{E4} observed that the transformed question stem projection does not form a neat cluster and does not align well with the target concept projection. It suggests insufficient learning of this relation.
In addition, there is a group of outliers at the top. Almost all of them have red color, which indicates a low model accuracy.
He wondered if it is the poor learning of \texttt{causes} relation that causes such high errors.
Afterward, \imp{E4} lassoed these instances to inspect their details further in the \sv{}.

\textbf{Subset Exploration}
(\imp{R1, R2, R3}) 
In the \sv{}, 
\imp{E4} noticed low model accuracy across all clusters, indicated by short green bars at the top of glyphs in \renfei{\autoref{fig:teaser}B}. 
Moreover, he spotted that question stem clusters have long blue bars compared to short green bars, implying that the model considers \cpn{} concepts in question stems but still fails to answer correctly. 
This strengthened \imp{E4}'s concerns over the model learning of the \texttt{cause} relation.
He clicked on the first question stem cluster (\renfei{\autoref{fig:teaser}B}) to explore its instances in the \iv{}.

\textbf{Instance Exploration and Editing}
(\imp{R1, R4}) 
\vis{
\imp{E4} first clicked the gray part of the accuracy bars to focus on the incorrect instances.
When scanning the model concepts (with stop words filtered) in the histogram, \imp{E4} found that the model usually attaches importance to words related to mood and emotion, such as ``happiness'', ``exhaustion'', and ``boredom''. 
Given the low model accuracy, \imp{E4} surmised that the model is not aware of the \texttt{causes} for human emotion.
To verify his thought, he explored and edited the detailed instances below. He found that most of these questions are very short, directly asking QC-TC relations. For example, as shown in \todo{\autoref{fig:teaser}C}, even after simplifying the original questions to ask straight at the relation between ``releasing built-up energy'' and ``wonderful'', the model still chose the wrong answer ``exhaustion''. He bookmarked these instances for model editing. 
}

\vis{\textbf{Model Editing} (\imp{R4}) 
\imp{E4} concluded that emotion-related \texttt{causes} is not sufficiently learned. And he applied model edits on the previously saved instances in the Model Editing Panel (\renfeivis{\autoref{fig:teaser}D}). He found that the editing accuracy is 100\% and model performance decrement is small (\ie, 71.01\%-70.93\%=0.08\%). He was satisfied with the edits.
}

\revvv{\subsubsection{Probe Model Limitations in Understanding Relation Contexts via Instance Exploration, Editing, and Querying}}
\label{subsec.case3}
\vspace{-5mm}
\begin{figure}[ht]
    \includegraphics[width=\columnwidth]{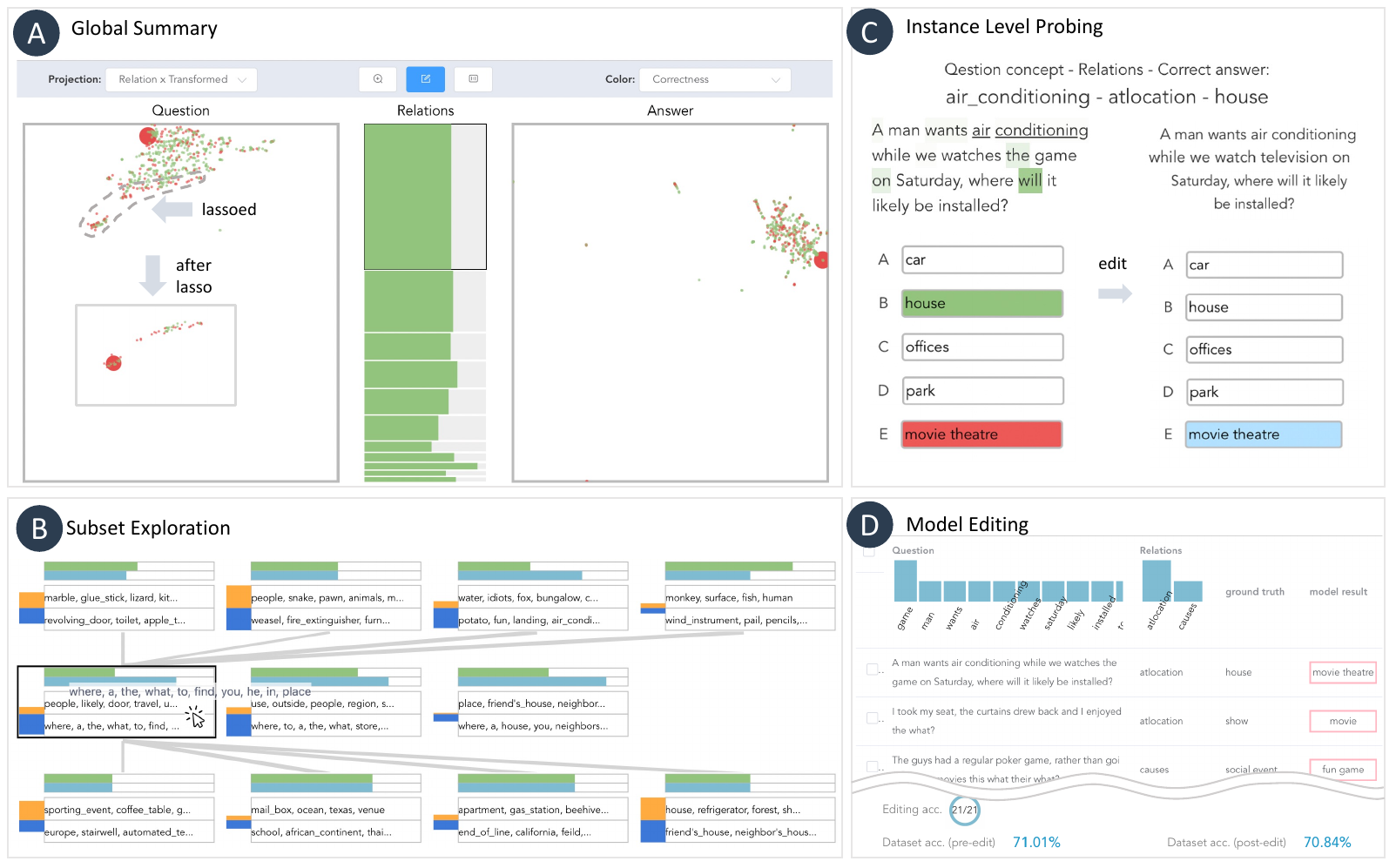}
    
    \caption{\renfeivis{The workflow \imp{E5} performed in \renfeivis{case two}: (A) \imp{E5} was interested in the border area and lassoed these points for further inspection. (B) \imp{E5} noticed a cluster with lower accuracy and clicked to inspect further. }}
    \label{fig:case2_new}
\vspace{-5mm}
\end{figure}


\textbf{Global Summary}
(\imp{R1, R2}) 
Another expert \imp{E5} used our system to investigate
under what circumstances the model might fail to use contexts for relational reasoning over concepts.
He first chose the largest relation group (\ie, \texttt{atlocation}, the first green bar) in the \gv{} (\renfei{\autoref{fig:case2_new}A}). 
And he found good correspondence between question stem and target concept clusters under the \textit{``Relation X Transformed''} projection scheme. 
It implies good learning of \texttt{atlocation} in general.
\imp{E5} wondered when the model might fail to reason about contexts. Then, he noticed that a group of dense red dots appear at the bottom left (\renfei{\autoref{fig:case2_new}A}). 
The accuracy of this group is low, and \imp{E5} decided to lasso the group for further exploration.

\textbf{Subset Exploration}
(\imp{R1, R2, R3}) 
In the \sv{}, \imp{E5} noticed that QSs fall into three clusters with varied accuracies (as suggested by the green bars) (\renfei{\autoref{fig:case2_new}B}).
Particularly, he was interested in the leftmost cluster since it has the lowest accuracy yet a similarly high question stem hit ratio (also indicated by tall dark blue bars at the left), compared to the other two.
As he hovered over the cluster glyph (\renfei{\autoref{fig:case2_new}B}), he discovered that the top model concepts are not so meaningful (\eg, a, the, to, you). 
He speculated that the model frequently relies on superficial information in question stems for answering the questions. 
\imp{E5} then clicked the cluster to explore the instances and model explanations. 


\textbf{Instance Manipulation}
(\imp{R1, R4}) 
In the \iv{}, \imp{E5} was curious about the incorrect instances and thus clicked the gray parts in the bar of model accuracy at the top to filter them.
When exploring the cases below, he found several interesting ones whose contexts associate with ``movie''.
For example, in one question (\renfei{\autoref{fig:case2_new}C}), 
\imp{E5} found that although ``air conditioning'' can also locate at ``movie theatre'', in this case, the model ignores the important context ``watch the game'' (without a green background), which normally happens in ``house''. 
Then, he further modified this instance to verify his finding.
Specifically, through several edits around ``watches'' in the original question (\renfei{\autoref{fig:case2_new}C}), the model still chooses ``movie theatre'' even though those contents such as television or live shows usually do not happen at ``movie theatre''. 
Therefore, \imp{E5} thought that the model attaches superficial information of ``watch'' to ``movie'' and does not understand the contexts.
In addition, other similar cases were observed where the model chooses ``movie'' without understanding what normally does not occur when watching movies, such as ``curtains drawing back'' or ``audiences clapping''.

\textbf{Instance query \& model editing} (\imp{R4}) 
\imp{E5} concluded that the model probably does not understand the contexts around ``movie'' well. He then located the related instances by searching keywords ``movie'' in the search input of \iv{}. He added those incorrect instances for model editing in the Model Editing Panel (\renfei{\autoref{fig:case2_new}D}). Finally, he saw that the editing accuracy is 100\% and the model maintains nearly the same performance as the original version (\ie, 70.84\% v.s. 71.01\%).


\subsection{User Study}\label{subsec.user_study}


We describe a user study that investigates how NLP experts utilize different components of \name{} to understand and diagnose models' commonsense reasoning capabilities. We also summarize their feedback on our system workflow and designs.

\subsubsection{Experiment Design}

\textbf{Participants}
We recruited 10 postgraduate students and alumni (eight males and two females, age: 20-30, referred as \imp{E4}-\imp{E13}) from the computer science department of a local university through emails and word-of-mouth. 
They had at least two years of experience in developing and evaluating natural language understanding models in academia or industry. 
None of the participants had prior involvement in our system's design or usage. Each participant received a cash compensation of \$13.

\textbf{User tasks}
Participants were required to use \name{} to analyze 
UnifiedQA~\cite{khashabi2022unifiedqa} on the CSQA validation set.
They needed to finish the following tasks: 1) gain an overview of model performance for different concepts and relations;
2) Find a relation of interest and assess the overall model learning of that relation;
3) Find a cluster of instances in the question stem/target concept scatter plots with large/small errors;
4) Summarize the model behavior on the cluster of instances;
5) Explore individual instances and infer if the model learns some commonsense to reason about concepts and their underlying relations.

\textbf{Procedures}
The whole study lasted about one hour. It started with a \xingbo{20-minute} tutorial, where we collected participants' demographics, asked for their permission to use their log data generated during the study, and introduced the background and the system usage. 
Afterward, participants could freely explore the system and get familiar with it (15 minutes). Then, they were asked to use our system to finish the aforementioned tasks. They were encouraged to speak out their hypotheses and findings about the model following a think-aloud protocol (20 minutes). 
During the task exploration, all their user interaction activities (\eg, lasso, clicking, and hovering), together with the timestamps, were automatically recorded.
Finally, participants needed to finish a questionnaire about system workflow, designs, and usability \revv{on a 5-point Likert scale}. And we collected their post-study feedback on the experience of using \name{}.

\subsubsection{Results and Analysis}

We report the analysis of user log data, the questionnaire, and participants' feedback.
For user logs, we extracted the frequency and duration of individual interactions (\eg, clicking) and aggregated them to derive the total usage frequency and duration of corresponding views.

\textbf{Model behavior contextualization and alignment}
Participants found it reasonable and helpful to use ConceptNet to contextualize the model's commonsense reasoning abilities on the CSQA dataset. 


Most participants agreed that \name{} helped them understand the data ($Mean_{Q1}=4.70$, $SD_{Q1}=0.67$), model performance  ($Mean_{Q2}=4.60$, $SD_{Q2}=0.70$) regarding different types of commonsense knowledge, and infer the model's implicit reasoning over concepts and their latent relations ($Mean_{Q3}=4.40$, $SD_{Q3}=0.70$). 
For example, 
\imp{E6} appreciated the intuitive assessment of overall relation learning by comparing transformed embeddings.
\imp{E7} mentioned, 
\hl{... getting overlap between the set of entities mapped in ConceptNet and model entities (concepts) can explain whether LMs are focusing on the right things.}
\imp{E8} added, \hl{Aggregating the concept hit ratio and model performance over the whole dataset and organizing them by relation is a great way to understand the model from a global scale.}
Moreover, participants were fairly confident in their findings ($Mean_{Q3}=4.40$, $SD_{Q3}=0.70$).
Besides, the extra context information provided by the ConceptNet helped develop hypotheses about model learning and probe the models' behavior on specific concepts and relations (\imp{E6, E9, E10}).

Nevertheless, participants also raised some concerns 
over using \cpn{}.
\imp{E10} said that sometimes the retrieved relations from ConceptNet are not guaranteed to be true reasoning paths for solving the questions.
Similarly, the extracted ConceptNet concepts were considered generic or not informative for some questions (\imp{E8}).

\begin{table}[t]
\centering
\caption{The frequencies and durations of user interactions.}
\resizebox{0.7\linewidth}{!}{
\begin{tabular}{ccc} 
\hline
View          & Frequency                          & Duration (s)                             \\ 
\hline
Global View   & 59.63 $\pm$ 32.32 & 468.48 $\pm$ 289.01  \\
Subset View   & 22.50 $\pm$ 17.12 & 66.82 $\pm$ 60.15    \\
Instance View & 33.38 $\pm$ 23.65 & 212.15 $\pm$ 204.15  \\
\hline
\end{tabular}
}
\label{tab:interaction-statistics}
\vspace{-5mm}
\end{table}


\begin{figure}[b]
\centering
    \vspace{-5mm}
    \includegraphics[width=\linewidth]{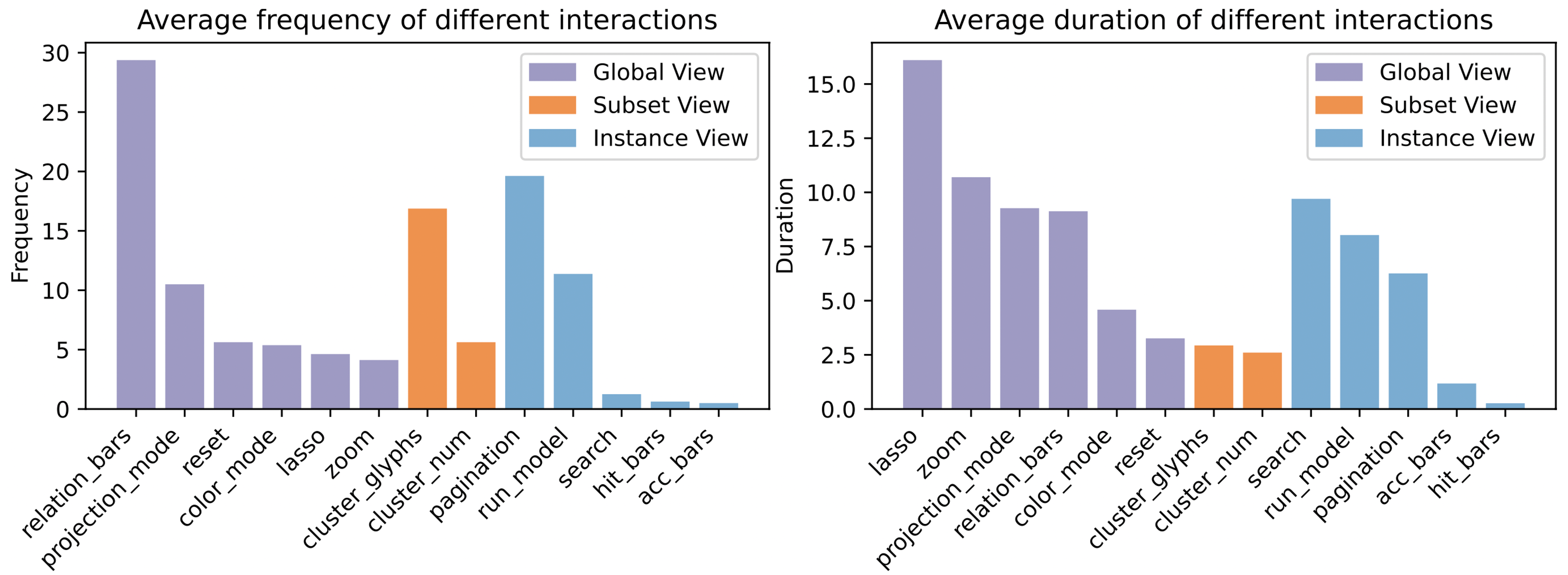}
    \vspace{-4mm}
    \caption{Average frequencies and durations of system interactions.}
    \label{fig:interaction_freq}
\end{figure}

\textbf{System usage analysis}
Participants thought that \name{} supports a more systematic and scalable analysis of model behavior, compared to conventional analysis of ad-hoc instances.
They appreciated the multi-level model explanations, especially the global understanding of model behavior.
As shown in \autoref{tab:interaction-statistics}, the \gv{} which provides an overview of model behavior clearly dominates the user interactions.
It takes up 51.62\% and 62.68\% of total system interaction frequency and duration respectively, exceeding those of the \sv{} and \iv{} by a large margin.
Moreover, as shown in \autoref{fig:interaction_freq},
participants showed great interest in examining model performance for different relations and projection modes. 
They spent much time selecting and exploring instances in the scatter plots (indicated by lasso and zoom interactions).
The \sv{} has the least user interactions regarding frequency and duration.
Participants usually used it to quickly preview the details of different cluster glyphs (indicated by ``cluster\_glyph'' interactions).
The \iv{} was considered to represent the traditional analysis of model behavior.
Participants generally used it to check different instances (indicated by ``pagination'' interaction) and probe the model (indicated by ``run\_model'' interaction). Also, they spent the longest time searching words and phrases in the \iv{} (indicated by ``search'' interaction).



\textbf{Patterns and insights}
Participants also reported many interesting findings about the model behavior and dataset issues.
For example, the model was found to rely on spurious correlations to solve many questions.
For the question \hl{Minerals can be obtained in what way for a person who avoids leafy greens? (answer: multivitamin)}, the model attached importance to ``minerals obtained'' and selects ``ore'', ignoring the important contexts ``person'' and ``leafy green''. 
Also in many instances, the model just focuses on some irrelevant words 
like ``what'' and ``at'' (\imp{E9, E11}).
\imp{E10} suggested that we can convert the questions into statements and check the model behavior change.
\hl{We can eliminate those word biases by using counterfactuals.} (\imp{E11}).

Besides, participants noticed that some CSQA questions are poorly designed. 
These questions have multiple plausible answers or typos that entirely invalidate the whole question instances.
For example, \imp{E11} mentioned, for the question ``Where is seaweed from?'', the model outputs ``sea'', while the true answer is defined as ``ocean''.
However, both ``sea'' and ``ocean'' seem correct.
And \hl{`what can a person with a what can do?' should’ve been `...with a watch...'} (\imp{E7}).

\textbf{Visual designs and interactions}
Participants generally agreed that our system is easy to use, but it required some effort to learn. 
They found the \iv{} to be the most intuitive, followed by the \gv{}. 
participants found the \iv{} the most helpful in diagnosing if the model uses proper information and learns relations between question and target concepts.
They thought that SHAP explanations and model probing complement each other to deepen the model understanding.
The \gv{} was thought useful in summarizing the learning of relations and concepts, though it was sometimes a bit hard to visually align dot clusters between the projection plots due to the embedding rotation effect and scarcity of instances. 
The \sv{} was considered the most difficult to understand.
But it was considered helpful to compare the model concepts and ConceptNet concepts across different subsets.
Their ratings and detailed feedback are in 
\referappendix{Suppl. E.2}.

\textbf{Suggestions for improvement}
\imp{E10} desired an automatic zoom-in function when lassoing dots in the \gv{}.
\imp{E11} suggested showing the neighboring dots when hovering over a dot. 
\imp{E8} proposed to summarize concepts that are not covered by both ConceptNet and the model.
\imp{E7} recommended integrating other knowledge bases (\eg, ATOMIC~\cite{atomic2019,atomic2020}) to broaden the commonsense coverage.
\section{Discussion}
\label{sec.discussion}



\subsection{Human-AI Alignment with Contextualization}
Considering that commonsense knowledge is implicit and not explicitly stated (in model input), many interpretability techniques~\cite{sharedinterest, lundberg2017unified}, which rely on existing model input and output, cannot explain models for commonsense reasoning tasks.
We introduce an external knowledge graph, \cpn{}, to characterize the commonsense knowledge in the model input with a group of concepts connected by different relations. 
This external knowledge is set as contextual information to align model behavior with human commonsense knowledge and reasoning.
Given the large space of commonsense knowledge, achieving human-AI alignment on commonsense reasoning tasks with additional contexts is challenging.
Our multi-level visualizations enable the exploration of model behavior on different concepts and their underlying relations in a scalable and systematic way.
 \vis{Moreover, visualizations produce actionable insights into what specific knowledge the model underperforms and guides the model probing and editing.
With pre-trained language models becoming so large and powerful (\eg, ChatGPT), it poses significant challenges to understand, diagnose, and adjust model behavior after deployment. \name{} presents \textit{``exploration-explanation-editing''} posthoc analysis pipeline to contextualize and align model behavior with users' expectations.
}

\subsection{Commonsense Knowledge Bases for Contextualization}
To reflect implicit commonsense knowledge in models, we use an external knowledge graph (\cpn{}) as the contextual reference of commonsense knowledge and align the model behavior with the \cpn{} knowledge. 
Given that ConceptNet is large and comprehensive with good generality and CSQA is built upon \cpn{}, it is reasonable and sufficient to use ConceptNet to cover the commonsense knowledge in CSQA.
\revv{Our quantitative evaluation with CSQA examples and qualitative users' feedback also show that most commonsense knowledge in CSQA questions can be grounded in ConceptNet, justifying its use in our study.}
However, \cpn{} mainly contains taxonomic, lexical, and physical commonsense. 
\revv{It has limitations in covering other commonsense knowledge, such as temporal and inferential commonsense, thus impacting the effectiveness of model behavior contextualization and visualization for understanding models' true reasoning capabilities in these knowledge areas.}
To support model analysis for more commonsense reasoning benchmarks, we can integrate diverse commonsense knowledge bases, such as ATOMIC~\cite{atomic2019,atomic2020}, GLUCOSE~\cite{glucose}, or large pretrained language models~\cite{shwartz2020unsupervised}, to contextualize model behavior.
\rev{It is worth noting that to ensure meaningful model contextualization,  we should choose a knowledge base that has sufficient and relevant coverage of the commonsense knowledge in the target benchmark dataset.
}
Besides knowledge graphs, other types of commonsense knowledge representations (\eg, arithmetic and logical operations) can be used to improve the expressiveness of model behavior contextualization.

\subsection{Generalizability and Scalability}
We showcase \name{} through a state-of-the-art QA language model (\ie, UnifiedQA) and CSQA dataset.
Our system can be immediately used to \textbf{explain any other language models} for commonsense question answering (CQA) since our explanations center around the model's input-output behavior with a model-agnostic approach.
\rev{Moreover, our system can be used for \textbf{other CQA datasets}. 
For example, Social IQA~\cite{socialiqa} is a multiple-choice QA dataset about social interactions in everyday events.
It is built upon ATOMIC~\cite{atomic2019,atomic2020}---a commonsense knowledge graph about the causes and effects of different events.
Therefore, our system can integrate ATOMIC to retrieve relevant causes or events for a given event extracted from questions in Social IQA to contextualize the model's social commonsense reasoning.}
\rev{Furthermore, our model contextualization method has the potential to support \textbf{other commonsense reasoning tasks}.
For instance, for visual question-answering (VQA) tasks, we can extract concepts (\eg, person, dog) in the images. Then, by combining the concepts in the images and text questions, we can utilize an external knowledge base to build a relevant knowledge graph that covers these concepts. The resulting knowledge graph can be used to contextualize models' reasoning.}
Our \textbf{multi-level visual designs} facilitate NLP model analysis for tasks like machine translation. The scatter plots of the \gv{} can summarize frequent associations and translation errors between source and target concepts. Aligning embeddings helps assess translation quality. Then, the \iv{} shows correlations between source and target sentences, enabling users to evaluate translation robustness.

\revv{Our approach faces \textbf{scalability} issues due to the computation cost of feature attribution methods, which can take several hours to compute SHAP for thousands of instances. 
To mitigate this impact, we have precomputed and integrated SHAP values into the system to enable seamless interactions for posthoc model analysis.
To further speed up the process, we can adopt faster feature attribution methods (\eg, CXplain~\cite{schwab2019cxplain}) and techniques like data sampling, caching, and parallel computing.}
Regarding visual designs, the links between cluster glyphs in the \sv{} can be cluttered when the cluster number exceeds five,
requiring horizontal scrolling to examine different clusters.


\subsection{Limitations and Future Work}
\name{} also has some limitations:
1) To extract relevant commonsense knowledge for answering a question, 
we build a sub-graph containing the
words in the question stem that are within two hops of the question concept and answer using ConceptNet. 
However, some concepts in the sub-graph may not so relevant for solving the question. 
Also, some important concepts could be connected with the question concept through multiple hops.
\rev{2) We perform n-gram tokenization to match the words in a question with the words in \cpn{}. However, this may exclude some longer phrases or sentences (in question stems and answers), which affects the relation extraction between the question and answer.}
3) To reflect the model's overall learning of relations, we apply the translation to the input embedding and align it with the output embedding.
However, it is also possible that after linear transformation, question and target concepts are not close to each other, but the models still capture the relations between question and target concepts through non-linear transformation. 
\revv{4) Model behavior probing may lead to incorrect model understanding~\cite{slack2021counterfactual}. To mitigate this, the system can improve the reliability of probing~\cite{voita-titov-2020-information,poyiadzi2020face} or integrate multiple explanation methods to cross-validate the model insights discovered by model probing. Moreover, larger-scale evaluation across different datasets and longer-term user studies with our experts can further validate the model understanding facilitated by our system.}

\revv{In the future, we can improve the system designs to handle more complex questions with multiple plausible answers and explanations, often influenced by diverse arguments and opinions.}
Moreover, we can improve the usability of \name{} by displaying prediction scores for answer choices in the \iv{} and enhancing our model editing methods for larger-scale editing.

\section{Conclusion}
\label{sec.conclusion}

We presented \name{} to help NLP experts to conduct a systematic and scalable analysis of the commonsense reasoning capabilities of language models.
It utilized an external commonsense knowledge base to contextualize and visualize 
the model behavior on different concepts and underlying relations from different levels of detail.
Users can interactively probe and edit the model behavior to improve the model's reasoning abilities in specific knowledge areas.
The user study with cases showed the effectiveness of \name{} for diagnosing what commonsense knowledge a language model learns.


\acknowledgments{%
The authors wish to thank anonymous reviewers for their valuable feedback. 
This research was supported in part by Hong Kong Theme-based Research Scheme grant T41-709/17N and a grant from MSRA. 
}

\bibliographystyle{abbrv-doi-hyperref}

\bibliography{main}

\begin{thebibliography}{10}

\bibitem{adadi2018xaisurvey}
A.~Adadi and M.~Berrada.
\newblock Peeking inside the black-box: A survey on explainable artificial
  intelligence (xai).
\newblock {\em IEEE Access}, 6:52138--52160, 2018.
  \href{https://doi.org/10.1109/ACCESS.2018.2870052}
{doi: {{%
10\hspace{.1pt}\discretionary{.}{%
}{.}\hspace{.4pt}1109\discretionary{/}{%
}{/}ACCESS\hspace{.1pt}\discretionary{.}{%
}{.}\hspace{.4pt}2018\hspace{.1pt}\discretionary{.}{%
}{.}\hspace{.4pt}2870052}}}


\bibitem{amershi2015modeltracker}
S.~Amershi, M.~Chickering, S.~M. Drucker, B.~Lee, P.~Simard, and J.~Suh.
\newblock Modeltracker: Redesigning performance analysis tools for machine
  learning.
\newblock In {\em Proc. CHI}, pp. 337--346. ACM, New York, 2015.
  \href{https://doi.org/10.1145/2702123.2702509}
{doi: {{%
10\hspace{.1pt}\discretionary{.}{%
}{.}\hspace{.4pt}1145\discretionary{/}{%
}{/}2702123\hspace{.1pt}\discretionary{.}{%
}{.}\hspace{.4pt}2702509}}}


\bibitem{ayoub2021combat}
J.~Ayoub, X.~J. Yang, and F.~Zhou.
\newblock Combat covid-19 infodemic using explainable natural language
  processing models.
\newblock {\em Inf. Process. Manage.}, 58(4):102569, 2021.
  \href{https://doi.org/10.1016/j.ipm.2021.102569}
{doi: {{%
10\hspace{.1pt}\discretionary{.}{%
}{.}\hspace{.4pt}1016\discretionary{/}{%
}{/}j\hspace{.1pt}\discretionary{.}{%
}{.}\hspace{.4pt}ipm\hspace{.1pt}\discretionary{.}{%
}{.}\hspace{.4pt}2021\hspace{.1pt}\discretionary{.}{%
}{.}\hspace{.4pt}102569}}}


\bibitem{xai_survey}
A.~{Barredo Arrieta}, N.~DÃ­az-RodrÃ­guez, J.~{Del Ser}, A.~Bennetot,
  S.~Tabik, A.~Barbado, S.~Garcia, S.~Gil-Lopez, D.~Molina, R.~Benjamins,
  R.~Chatila, and F.~Herrera.
\newblock Explainable artificial intelligence (xai): Concepts, taxonomies,
  opportunities and challenges toward responsible ai.
\newblock {\em Inf. Fusion}, 58:82--115, 2020.
  \href{https://doi.org/10.1016/j.inffus.2019.12.012}
{doi: {{%
10\hspace{.1pt}\discretionary{.}{%
}{.}\hspace{.4pt}1016\discretionary{/}{%
}{/}j\hspace{.1pt}\discretionary{.}{%
}{.}\hspace{.4pt}inffus\hspace{.1pt}\discretionary{.}{%
}{.}\hspace{.4pt}2019\hspace{.1pt}\discretionary{.}{%
}{.}\hspace{.4pt}12\hspace{.1pt}\discretionary{.}{%
}{.}\hspace{.4pt}012}}}


\bibitem{bhagavatula2019abductive}
C.~Bhagavatula, R.~L. Bras, C.~Malaviya, K.~Sakaguchi, A.~Holtzman, H.~Rashkin,
  D.~Downey, S.~W.-t. Yih, and Y.~Choi.
\newblock Abductive commonsense reasoning.
\newblock In {\em ICLR}, 2020.

\bibitem{bian2023chatgpt}
N.~Bian, X.~Han, L.~Sun, H.~Lin, Y.~Lu, and B.~He.
\newblock Chatgpt is a knowledgeable but inexperienced solver: An investigation
  of commonsense problem in large language models.
\newblock {\em arXiv preprint arXiv:2303.16421}, 2023.
  \href{https://doi.org/10.48550/arXiv.2303.16421}
{doi: {{%
10\hspace{.1pt}\discretionary{.}{%
}{.}\hspace{.4pt}48550\discretionary{/}{%
}{/}arXiv\hspace{.1pt}\discretionary{.}{%
}{.}\hspace{.4pt}2303\hspace{.1pt}\discretionary{.}{%
}{.}\hspace{.4pt}16421}}}


\bibitem{bisk2020piqa}
Y.~Bisk, R.~Zellers, R.~L. Bras, J.~Gao, and Y.~Choi.
\newblock Piqa: Reasoning about physical commonsense in natural language.
\newblock In {\em AAAI}, 2020. \href{https://doi.org/10.1609/aaai.v34i05.6239}
{doi: {{%
10\hspace{.1pt}\discretionary{.}{%
}{.}\hspace{.4pt}1609\discretionary{/}{%
}{/}aaai\hspace{.1pt}\discretionary{.}{%
}{.}\hspace{.4pt}v34i05\hspace{.1pt}\discretionary{.}{%
}{.}\hspace{.4pt}6239}}}


\bibitem{sharedinterest}
A.~Boggust, B.~Hoover, A.~Satyanarayan, and H.~Strobelt.
\newblock Shared interest: Measuring human-ai alignment to identify recurring
  patterns in model behavior.
\newblock In {\em Proc. CHI}, pp. 1--17. ACM, New York, 2022.
  \href{https://doi.org/10.1145/3491102.3501965}
{doi: {{%
10\hspace{.1pt}\discretionary{.}{%
}{.}\hspace{.4pt}1145\discretionary{/}{%
}{/}3491102\hspace{.1pt}\discretionary{.}{%
}{.}\hspace{.4pt}3501965}}}


\bibitem{boratko2020protoqa}
M.~Boratko, X.~L. Li, R.~Das, T.~O'Gorman, D.~Le, and A.~McCallum.
\newblock Protoqa: A question answering dataset for prototypical common-sense
  reasoning.
\newblock In {\em Proc. EMNLP}, pp. 1122--1136. ACL, Online, 2020.
  \href{https://doi.org/10.18653/v1/2020.emnlp-main.85}
{doi: {{%
10\hspace{.1pt}\discretionary{.}{%
}{.}\hspace{.4pt}18653\discretionary{/}{%
}{/}v1\discretionary{/}{%
}{/}2020\hspace{.1pt}\discretionary{.}{%
}{.}\hspace{.4pt}emnlp\discretionary{%
}{-}{-}main\hspace{.1pt}\discretionary{.}{%
}{.}\hspace{.4pt}85}}}


\bibitem{bordes2013translating}
A.~Bordes, N.~Usunier, A.~Garcia-Duran, J.~Weston, and O.~Yakhnenko.
\newblock Translating embeddings for modeling multi-relational data.
\newblock In {\em NeurIPS}, vol.~26, pp. 2787--2795, 2013.

\bibitem{gpt3}
T.~B. Brown, B.~Mann, N.~Ryder, M.~Subbiah, J.~Kaplan, P.~Dhariwal,
  A.~Neelakantan, P.~Shyam, G.~Sastry, A.~Askell, S.~Agarwal, A.~Herbert-Voss,
  G.~Krueger, T.~Henighan, R.~Child, A.~Ramesh, D.~M. Ziegler, J.~Wu,
  C.~Winter, C.~Hesse, M.~Chen, E.~Sigler, M.~Litwin, S.~Gray, B.~Chess,
  J.~Clark, C.~Berner, S.~McCandlish, A.~Radford, I.~Sutskever, and D.~Amodei.
\newblock Language models are few-shot learners.
\newblock In {\em NeurIPS}, vol.~33, pp. 1877--1901, 2020.

\bibitem{chen2020generating}
H.~Chen, G.~Zheng, and Y.~Ji.
\newblock Generating hierarchical explanations on text classification via
  feature interaction detection.
\newblock In {\em Proc. ACL}, pp. 5578--5593. ACL, Online, 2020.
  \href{https://doi.org/10.18653/v1/2020.acl-main.494}
{doi: {{%
10\hspace{.1pt}\discretionary{.}{%
}{.}\hspace{.4pt}18653\discretionary{/}{%
}{/}v1\discretionary{/}{%
}{/}2020\hspace{.1pt}\discretionary{.}{%
}{.}\hspace{.4pt}acl\discretionary{%
}{-}{-}main\hspace{.1pt}\discretionary{.}{%
}{.}\hspace{.4pt}494}}}


\bibitem{cui2020commonsense}
L.~Cui, S.~Cheng, Y.~Wu, and Y.~Zhang.
\newblock On commonsense cues in bert for solving commonsense tasks.
\newblock In {\em Findings of ACL: ACL/IJCNLP}, pp. 683--693. ACL, online,
  2021. \href{https://doi.org/10.18653/v1/2021.findings-acl.61}
{doi: {{%
10\hspace{.1pt}\discretionary{.}{%
}{.}\hspace{.4pt}18653\discretionary{/}{%
}{/}v1\discretionary{/}{%
}{/}2021\hspace{.1pt}\discretionary{.}{%
}{.}\hspace{.4pt}findings\discretionary{%
}{-}{-}acl\hspace{.1pt}\discretionary{.}{%
}{.}\hspace{.4pt}61}}}


\bibitem{dai-etal-2022-knowledge}
D.~Dai, L.~Dong, Y.~Hao, Z.~Sui, B.~Chang, and F.~Wei.
\newblock Knowledge neurons in pretrained transformers.
\newblock In {\em Proc. ACL}, pp. 8493--8502. ACL, Dublin, Ireland, 2022.
  \href{https://doi.org/10.18653/v1/2022.acl-long.581}
{doi: {{%
10\hspace{.1pt}\discretionary{.}{%
}{.}\hspace{.4pt}18653\discretionary{/}{%
}{/}v1\discretionary{/}{%
}{/}2022\hspace{.1pt}\discretionary{.}{%
}{.}\hspace{.4pt}acl\discretionary{%
}{-}{-}long\hspace{.1pt}\discretionary{.}{%
}{.}\hspace{.4pt}581}}}


\bibitem{de2021editing}
N.~De~Cao, W.~Aziz, and I.~Titov.
\newblock Editing factual knowledge in language models.
\newblock In {\em Proc. EMNLP}, pp. 6491--6506. ACL, Online and Punta Cana,
  Dominican Republic, 2021.
  \href{https://doi.org/10.18653/v1/2021.emnlp-main.522}
{doi: {{%
10\hspace{.1pt}\discretionary{.}{%
}{.}\hspace{.4pt}18653\discretionary{/}{%
}{/}v1\discretionary{/}{%
}{/}2021\hspace{.1pt}\discretionary{.}{%
}{.}\hspace{.4pt}emnlp\discretionary{%
}{-}{-}main\hspace{.1pt}\discretionary{.}{%
}{.}\hspace{.4pt}522}}}


\bibitem{dinu2014improving}
G.~Dinu and M.~Baroni.
\newblock Improving zero-shot learning by mitigating the hubness problem.
\newblock In {\em ICLR}, 2015.

\bibitem{mediators}
N.~Feldhus, A.~M. Ravichandran, and S.~M{\"o}ller.
\newblock Mediators: Conversational agents explaining nlp model behavior.
\newblock {\em arXiv preprint arXiv:2206.06029}, 2022.
  \href{https://doi.org/10.48550/arXiv.2206.06029}
{doi: {{%
10\hspace{.1pt}\discretionary{.}{%
}{.}\hspace{.4pt}48550\discretionary{/}{%
}{/}arXiv\hspace{.1pt}\discretionary{.}{%
}{.}\hspace{.4pt}2206\hspace{.1pt}\discretionary{.}{%
}{.}\hspace{.4pt}06029}}}


\bibitem{feng2020scalable}
Y.~Feng, X.~Chen, B.~Y. Lin, P.~Wang, J.~Yan, and X.~Ren.
\newblock Scalable multi-hop relational reasoning for knowledge-aware question
  answering.
\newblock In {\em Proc. EMNLP}, pp. 1295--1309. ACL, Online, 2020.
  \href{https://doi.org/10.18653/v1/2020.emnlp-main.99}
{doi: {{%
10\hspace{.1pt}\discretionary{.}{%
}{.}\hspace{.4pt}18653\discretionary{/}{%
}{/}v1\discretionary{/}{%
}{/}2020\hspace{.1pt}\discretionary{.}{%
}{.}\hspace{.4pt}emnlp\discretionary{%
}{-}{-}main\hspace{.1pt}\discretionary{.}{%
}{.}\hspace{.4pt}99}}}


\bibitem{feng2023xnli}
Y.~Feng, X.~Wang, B.~Pan, K.~K. Wong, Y.~Ren, S.~Liu, Z.~Yan, Y.~Ma, H.~Qu, and
  W.~Chen.
\newblock Xnli: Explaining and diagnosing nli-based visual data analysis.
\newblock {\em IEEE Trans. Visual Comput. Graphics}, 2023.
  \href{https://doi.org/10.1109/TVCG.2023.3240003}
{doi: {{%
10\hspace{.1pt}\discretionary{.}{%
}{.}\hspace{.4pt}1109\discretionary{/}{%
}{/}TVCG\hspace{.1pt}\discretionary{.}{%
}{.}\hspace{.4pt}2023\hspace{.1pt}\discretionary{.}{%
}{.}\hspace{.4pt}3240003}}}


\bibitem{hoover2019exbert}
B.~Hoover, H.~Strobelt, and S.~Gehrmann.
\newblock ex{BERT}: {A} visual analysis tool to explore learned representations
  in transformer models.
\newblock In {\em Proc. ACL: System Demonstrations}, pp. 187--196, 2020.
  \href{https://doi.org/10.18653/v1/2020.acl-demos.22}
{doi: {{%
10\hspace{.1pt}\discretionary{.}{%
}{.}\hspace{.4pt}18653\discretionary{/}{%
}{/}v1\discretionary{/}{%
}{/}2020\hspace{.1pt}\discretionary{.}{%
}{.}\hspace{.4pt}acl\discretionary{%
}{-}{-}demos\hspace{.1pt}\discretionary{.}{%
}{.}\hspace{.4pt}22}}}


\bibitem{huang2019cosmos}
L.~Huang, R.~L. Bras, C.~Bhagavatula, and Y.~Choi.
\newblock Cosmos qa: Machine reading comprehension with contextual commonsense
  reasoning.
\newblock In {\em Proc. EMNLP}, pp. 2391--2401. ACL, Hong Kong, 2019.
  \href{https://doi.org/10.18653/v1/D19-1243}
{doi: {{%
10\hspace{.1pt}\discretionary{.}{%
}{.}\hspace{.4pt}18653\discretionary{/}{%
}{/}v1\discretionary{/}{%
}{/}D19\discretionary{%
}{-}{-}1243}}}


\bibitem{atomic2020}
J.~D. Hwang, C.~Bhagavatula, R.~L. Bras, J.~Da, K.~Sakaguchi, A.~Bosselut, and
  Y.~Choi.
\newblock Comet-atomic 2020: On symbolic and neural commonsense knowledge
  graphs.
\newblock In {\em AAAI}, pp. 6384--6392, 2021.
  \href{http://dx.doi.org/10.1609/aaai.v35i7.16792}
{doi: {{%
10\hspace{.1pt}\discretionary{.}{%
}{.}\hspace{.4pt}1609\discretionary{/}{%
}{/}aaai\hspace{.1pt}\discretionary{.}{%
}{.}\hspace{.4pt}v35i7\hspace{.1pt}\discretionary{.}{%
}{.}\hspace{.4pt}16792}}}


\bibitem{jin2023shortcutlens}
Z.~Jin, X.~Wang, F.~Cheng, C.~Sun, Q.~Liu, and H.~Qu.
\newblock Shortcutlens: A visual analytics approach for exploring shortcuts in
  natural language understanding dataset.
\newblock {\em IEEE Trans. Visual Comput. Graphics}, 2023.
  \href{https://doi.org/10.1109/TVCG.2023.3236380}
{doi: {{%
10\hspace{.1pt}\discretionary{.}{%
}{.}\hspace{.4pt}1109\discretionary{/}{%
}{/}TVCG\hspace{.1pt}\discretionary{.}{%
}{.}\hspace{.4pt}2023\hspace{.1pt}\discretionary{.}{%
}{.}\hspace{.4pt}3236380}}}


\bibitem{jin2022gnnlens}
Z.~Jin, Y.~Wang, Q.~Wang, Y.~Ming, T.~Ma, and H.~Qu.
\newblock Gnnlens: A visual analytics approach for prediction error diagnosis
  of graph neural networks.
\newblock {\em IEEE Trans. Visual Comput. Graphics}, 2022.
  \href{https://doi.org/10.1109/TVCG.2022.3148107}
{doi: {{%
10\hspace{.1pt}\discretionary{.}{%
}{.}\hspace{.4pt}1109\discretionary{/}{%
}{/}TVCG\hspace{.1pt}\discretionary{.}{%
}{.}\hspace{.4pt}2022\hspace{.1pt}\discretionary{.}{%
}{.}\hspace{.4pt}3148107}}}


\bibitem{kaushik2019learning}
D.~Kaushik, E.~Hovy, and Z.~C. Lipton.
\newblock Learning the difference that makes a difference with
  counterfactually-augmented data.
\newblock In {\em ICLR}, 2020.

\bibitem{khashabi2022unifiedqa}
D.~Khashabi, Y.~Kordi, and H.~Hajishirzi.
\newblock Unifiedqa-v2: Stronger generalization via broader cross-format
  training.
\newblock {\em arXiv preprint arXiv:2202.12359}, 2022.
  \href{https://doi.org/10.48550/arXiv.2202.12359}
{doi: {{%
10\hspace{.1pt}\discretionary{.}{%
}{.}\hspace{.4pt}48550\discretionary{/}{%
}{/}arXiv\hspace{.1pt}\discretionary{.}{%
}{.}\hspace{.4pt}2202\hspace{.1pt}\discretionary{.}{%
}{.}\hspace{.4pt}12359}}}


\bibitem{2020unifiedqa}
D.~Khashabi, S.~Min, T.~Khot, A.~Sabhwaral, O.~Tafjord, P.~Clark, and
  H.~Hajishirzi.
\newblock Unifiedqa: Crossing format boundaries with a single qa system.
\newblock In {\em Findings of ACL: EMNLP}, pp. 1896--1907. ACL, online, 2020.
  \href{https://doi.org/10.18653/v1/2020.findings-emnlp.171}
{doi: {{%
10\hspace{.1pt}\discretionary{.}{%
}{.}\hspace{.4pt}18653\discretionary{/}{%
}{/}v1\discretionary{/}{%
}{/}2020\hspace{.1pt}\discretionary{.}{%
}{.}\hspace{.4pt}findings\discretionary{%
}{-}{-}emnlp\hspace{.1pt}\discretionary{.}{%
}{.}\hspace{.4pt}171}}}


\bibitem{li-etal-2022-systematic}
X.~L. Li, A.~Kuncoro, J.~Hoffmann, C.~de~Masson~d{'}Autume, P.~Blunsom, and
  A.~Nematzadeh.
\newblock A systematic investigation of commonsense knowledge in large language
  models.
\newblock In {\em Proc. EMNLP}, pp. 11838--11855. ACL, Abu Dhabi, 2022.

\bibitem{li2022unified}
Z.~Li, X.~Wang, W.~Yang, J.~Wu, Z.~Zhang, Z.~Liu, M.~Sun, H.~Zhang, and S.~Liu.
\newblock A unified understanding of deep nlp models for text classification.
\newblock {\em IEEE Trans. Visual Comput. Graphics}, 28(12):4980--4994, 2022.
  \href{https://doi.org/10.1109/TVCG.2022.3184186}
{doi: {{%
10\hspace{.1pt}\discretionary{.}{%
}{.}\hspace{.4pt}1109\discretionary{/}{%
}{/}TVCG\hspace{.1pt}\discretionary{.}{%
}{.}\hspace{.4pt}2022\hspace{.1pt}\discretionary{.}{%
}{.}\hspace{.4pt}3184186}}}


\bibitem{liang2022multiviz}
P.~P. Liang, Y.~Lyu, G.~Chhablani, N.~Jain, Z.~Deng, X.~Wang, L.-P. Morency,
  and R.~Salakhutdinov.
\newblock Multiviz: Towards visualizing and understanding multimodal models.
\newblock In {\em ICLR}, 2022.

\bibitem{lin2019kagnet}
B.~Y. Lin, X.~Chen, J.~Chen, and X.~Ren.
\newblock Kagnet: Knowledge-aware graph networks for commonsense reasoning.
\newblock In {\em Proc. EMNLP}, pp. 2829--2839. ACL, Hong Kong, 2019.
  \href{https://doi.org/10.18653/v1/D19-1282}
{doi: {{%
10\hspace{.1pt}\discretionary{.}{%
}{.}\hspace{.4pt}18653\discretionary{/}{%
}{/}v1\discretionary{/}{%
}{/}D19\discretionary{%
}{-}{-}1282}}}


\bibitem{lin2020differentiable}
B.~Y. Lin, H.~Sun, B.~Dhingra, M.~Zaheer, X.~Ren, and W.~W. Cohen.
\newblock Differentiable open-ended commonsense reasoning.
\newblock In {\em Proc. NAACL}, pp. 4611--4625. ACL, Online, 2021.
  \href{https://doi.org/10.18653/v1/2021.naacl-main.366}
{doi: {{%
10\hspace{.1pt}\discretionary{.}{%
}{.}\hspace{.4pt}18653\discretionary{/}{%
}{/}v1\discretionary{/}{%
}{/}2021\hspace{.1pt}\discretionary{.}{%
}{.}\hspace{.4pt}naacl\discretionary{%
}{-}{-}main\hspace{.1pt}\discretionary{.}{%
}{.}\hspace{.4pt}366}}}


\bibitem{lin2021riddlesense}
B.~Y. Lin, Z.~Wu, Y.~Yang, D.-H. Lee, and X.~Ren.
\newblock Riddlesense: Reasoning about riddle questions featuring linguistic
  creativity and commonsense knowledge.
\newblock In {\em Findings of ACL: ACL/IJCNLP}, pp. 1504--1515. ACL, Online,
  2021. \href{https://doi.org/10.18653/v1/2021.findings-acl.131}
{doi: {{%
10\hspace{.1pt}\discretionary{.}{%
}{.}\hspace{.4pt}18653\discretionary{/}{%
}{/}v1\discretionary{/}{%
}{/}2021\hspace{.1pt}\discretionary{.}{%
}{.}\hspace{.4pt}findings\discretionary{%
}{-}{-}acl\hspace{.1pt}\discretionary{.}{%
}{.}\hspace{.4pt}131}}}


\bibitem{liu2021generated}
J.~Liu, A.~Liu, X.~Lu, S.~Welleck, P.~West, R.~L. Bras, Y.~Choi, and
  H.~Hajishirzi.
\newblock Generated knowledge prompting for commonsense reasoning.
\newblock In {\em Proc. ACL}, pp. 3154--3169. ACL, Dublin, Ireland, 2022.
  \href{https://doi.org/10.18653/v1/2022.acl-long.225}
{doi: {{%
10\hspace{.1pt}\discretionary{.}{%
}{.}\hspace{.4pt}18653\discretionary{/}{%
}{/}v1\discretionary{/}{%
}{/}2022\hspace{.1pt}\discretionary{.}{%
}{.}\hspace{.4pt}acl\discretionary{%
}{-}{-}long\hspace{.1pt}\discretionary{.}{%
}{.}\hspace{.4pt}225}}}


\bibitem{unicorn}
N.~Lourie, R.~L. Bras, C.~Bhagavatula, and Y.~Choi.
\newblock Unicorn on rainbow: A universal commonsense reasoning model on a new
  multitask benchmark.
\newblock In {\em AAAI}, pp. 13480--13488, 2021.
  \href{https://doi.org/10.1609/aaai.v35i15.17590}
{doi: {{%
10\hspace{.1pt}\discretionary{.}{%
}{.}\hspace{.4pt}1609\discretionary{/}{%
}{/}aaai\hspace{.1pt}\discretionary{.}{%
}{.}\hspace{.4pt}v35i15\hspace{.1pt}\discretionary{.}{%
}{.}\hspace{.4pt}17590}}}


\bibitem{lundberg2017unified}
S.~M. Lundberg and S.-I. Lee.
\newblock A unified approach to interpreting model predictions.
\newblock In {\em NeurIPS}, vol.~30, pp. 4765--4774, 2017.

\bibitem{ma2021knowledge}
K.~Ma, F.~Ilievski, J.~Francis, Y.~Bisk, E.~Nyberg, and A.~Oltramari.
\newblock Knowledge-driven data construction for zero-shot evaluation in
  commonsense question answering.
\newblock In {\em AAAI}, vol.~35, pp. 13507--13515, 2021.
  \href{https://doi.org/10.1609/aaai.v35i15.17593}
{doi: {{%
10\hspace{.1pt}\discretionary{.}{%
}{.}\hspace{.4pt}1609\discretionary{/}{%
}{/}aaai\hspace{.1pt}\discretionary{.}{%
}{.}\hspace{.4pt}v35i15\hspace{.1pt}\discretionary{.}{%
}{.}\hspace{.4pt}17593}}}


\bibitem{manning1999foundations}
C.~Manning and H.~Schutze.
\newblock {\em Foundations of statistical natural language processing}.
\newblock MIT press, 1999.

\bibitem{mcinnes2018umap}
L.~McInnes, J.~Healy, and J.~Melville.
\newblock Umap: Uniform manifold approximation and projection for dimension
  reduction.
\newblock {\em arXiv preprint arXiv:1802.03426}, 2018.
  \href{https://doi.org/10.48550/arXiv.1802.03426}
{doi: {{%
10\hspace{.1pt}\discretionary{.}{%
}{.}\hspace{.4pt}48550\discretionary{/}{%
}{/}arXiv\hspace{.1pt}\discretionary{.}{%
}{.}\hspace{.4pt}1802\hspace{.1pt}\discretionary{.}{%
}{.}\hspace{.4pt}03426}}}


\bibitem{meng2022locating}
K.~Meng, D.~Bau, A.~Andonian, and Y.~Belinkov.
\newblock Locating and editing factual knowledge in gpt.
\newblock In {\em ICLR}, 2022.

\bibitem{meng2022mass}
K.~Meng, A.~S. Sharma, A.~Andonian, Y.~Belinkov, and D.~Bau.
\newblock Mass-editing memory in a transformer.
\newblock In {\em ICLR}, 2023.

\bibitem{ming2017understanding}
Y.~Ming, S.~Cao, R.~Zhang, Z.~Li, Y.~Chen, Y.~Song, and H.~Qu.
\newblock Understanding hidden memories of recurrent neural networks.
\newblock In {\em Proc. VAST}, pp. 13--24. IEEE, 2017.
  \href{https://doi.org/10.1109/VAST.2017.8585721}
{doi: {{%
10\hspace{.1pt}\discretionary{.}{%
}{.}\hspace{.4pt}1109\discretionary{/}{%
}{/}VAST\hspace{.1pt}\discretionary{.}{%
}{.}\hspace{.4pt}2017\hspace{.1pt}\discretionary{.}{%
}{.}\hspace{.4pt}8585721}}}


\bibitem{mitchell2022fast}
E.~Mitchell, C.~Lin, A.~Bosselut, C.~Finn, and C.~D. Manning.
\newblock Fast model editing at scale.
\newblock In {\em ICLR}, 2022.

\bibitem{glucose}
N.~Mostafazadeh, A.~Kalyanpur, L.~Moon, D.~Buchanan, L.~Berkowitz, O.~Biran,
  and J.~Chu-Carroll.
\newblock {GLUCOSE}: {G}enera{L}ized and {CO}ntextualized story explanations.
\newblock In {\em Proc. EMNLP}, pp. 4569--4586. ACL, Online, 2020.
  \href{https://doi.org/10.18653/v1/2020.emnlp-main.370}
{doi: {{%
10\hspace{.1pt}\discretionary{.}{%
}{.}\hspace{.4pt}18653\discretionary{/}{%
}{/}v1\discretionary{/}{%
}{/}2020\hspace{.1pt}\discretionary{.}{%
}{.}\hspace{.4pt}emnlp\discretionary{%
}{-}{-}main\hspace{.1pt}\discretionary{.}{%
}{.}\hspace{.4pt}370}}}


\bibitem{DBLP:journals/corr/abs-2203-02155}
L.~Ouyang, J.~Wu, X.~Jiang, D.~Almeida, C.~L. Wainwright, P.~Mishkin, C.~Zhang,
  S.~Agarwal, K.~Slama, A.~Ray, J.~Schulman, J.~Hilton, F.~Kelton, L.~Miller,
  M.~Simens, A.~Askell, P.~Welinder, P.~F. Christiano, J.~Leike, and R.~Lowe.
\newblock Training language models to follow instructions with human feedback.
\newblock In {\em NeurIPS}, vol.~35, pp. 27730--27744, 2022.

\bibitem{petroni2019language}
F.~Petroni, T.~Rockt{\"a}schel, P.~Lewis, A.~Bakhtin, Y.~Wu, A.~H. Miller, and
  S.~Riedel.
\newblock Language models as knowledge bases?
\newblock In {\em Proc. EMNLP}, pp. 2463--2473. ACL, Hong Kong, 2019.
  \href{https://doi.org/10.18653/v1/D19-1250}
{doi: {{%
10\hspace{.1pt}\discretionary{.}{%
}{.}\hspace{.4pt}18653\discretionary{/}{%
}{/}v1\discretionary{/}{%
}{/}D19\discretionary{%
}{-}{-}1250}}}


\bibitem{poyiadzi2020face}
R.~Poyiadzi, K.~Sokol, R.~Santos-Rodriguez, T.~De~Bie, and P.~Flach.
\newblock Face: Feasible and actionable counterfactual explanations.
\newblock In {\em Proc. AIES}, pp. 344--350, 2020.
  \href{https://doi.org/10.1145/3375627.3375850}
{doi: {{%
10\hspace{.1pt}\discretionary{.}{%
}{.}\hspace{.4pt}1145\discretionary{/}{%
}{/}3375627\hspace{.1pt}\discretionary{.}{%
}{.}\hspace{.4pt}3375850}}}


\bibitem{t5model}
C.~Raffel, N.~Shazeer, A.~Roberts, K.~Lee, S.~Narang, M.~Matena, Y.~Zhou,
  W.~Li, and P.~J. Liu.
\newblock Exploring the limits of transfer learning with a unified text-to-text
  transformer.
\newblock {\em J. Mach. Learn. Res.}, 21(140):1--67, 2020.

\bibitem{rajani2019explain}
N.~F. Rajani, B.~McCann, C.~Xiong, and R.~Socher.
\newblock Explain yourself! leveraging language models for commonsense
  reasoning.
\newblock In {\em Proc. ACL}, pp. 4932--4942. ACL, Florence, Italy, 2019.
  \href{https://doi.org/10.18653/v1/p19-1487}
{doi: {{%
10\hspace{.1pt}\discretionary{.}{%
}{.}\hspace{.4pt}18653\discretionary{/}{%
}{/}v1\discretionary{/}{%
}{/}p19\discretionary{%
}{-}{-}1487}}}


\bibitem{reif2019visualizing}
E.~Reif, A.~Yuan, M.~Wattenberg, F.~B. Viegas, A.~Coenen, A.~Pearce, and
  B.~Kim.
\newblock Visualizing and measuring the geometry of bert.
\newblock In {\em NeurIPS}, vol.~32, pp. 8592--8600, 2019.

\bibitem{ribeiro2016should}
M.~T. Ribeiro, S.~Singh, and C.~Guestrin.
\newblock "{Why} should i trust you?": Explaining the predictions of any
  classifier.
\newblock In {\em Proc. KDD}, pp. 1135--1144. ACM, New York, 2016.
  \href{https://doi.org/10.1145/2939672.2939778}
{doi: {{%
10\hspace{.1pt}\discretionary{.}{%
}{.}\hspace{.4pt}1145\discretionary{/}{%
}{/}2939672\hspace{.1pt}\discretionary{.}{%
}{.}\hspace{.4pt}2939778}}}


\bibitem{sakaguchi2021winogrande}
K.~Sakaguchi, R.~L. Bras, C.~Bhagavatula, and Y.~Choi.
\newblock Winogrande: An adversarial winograd schema challenge at scale.
\newblock {\em Communications of the ACM}, 64(9):99--106, 2021.
  \href{https://doi.org/10.1145/3474381}
{doi: {{%
10\hspace{.1pt}\discretionary{.}{%
}{.}\hspace{.4pt}1145\discretionary{/}{%
}{/}3474381}}}


\bibitem{atomic2019}
M.~Sap, R.~Le~Bras, E.~Allaway, C.~Bhagavatula, N.~Lourie, H.~Rashkin, B.~Roof,
  N.~A. Smith, and Y.~Choi.
\newblock Atomic: An atlas of machine commonsense for if-then reasoning.
\newblock In {\em AAAI}, vol.~33, pp. 3027--3035, 2019.
  \href{https://doi.org/10.1609/aaai.v33i01.33013027}
{doi: {{%
10\hspace{.1pt}\discretionary{.}{%
}{.}\hspace{.4pt}1609\discretionary{/}{%
}{/}aaai\hspace{.1pt}\discretionary{.}{%
}{.}\hspace{.4pt}v33i01\hspace{.1pt}\discretionary{.}{%
}{.}\hspace{.4pt}33013027}}}


\bibitem{socialiqa}
M.~Sap, H.~Rashkin, D.~Chen, R.~L. Bras, and Y.~Choi.
\newblock Social iqa: Commonsense reasoning about social interactions.
\newblock In {\em Proc. EMNLP}, pp. 4462--4472. ACL, Hong Kong, 2019.
  \href{https://doi.org/10.18653/v1/D19-1454}
{doi: {{%
10\hspace{.1pt}\discretionary{.}{%
}{.}\hspace{.4pt}18653\discretionary{/}{%
}{/}v1\discretionary{/}{%
}{/}D19\discretionary{%
}{-}{-}1454}}}


\bibitem{schwab2019cxplain}
P.~Schwab and W.~Karlen.
\newblock Cxplain: Causal explanations for model interpretation under
  uncertainty.
\newblock In {\em NeurIPS}, vol.~32, pp. 10220--10230, 2019.

\bibitem{shwartz2020unsupervised}
V.~Shwartz, P.~West, R.~L. Bras, C.~Bhagavatula, and Y.~Choi.
\newblock Unsupervised commonsense question answering with self-talk.
\newblock In {\em Proc. EMNLP}, pp. 4615--4629. ACL, Online, 2020.
  \href{https://doi.org/10.18653/v1/2020.emnlp-main.373}
{doi: {{%
10\hspace{.1pt}\discretionary{.}{%
}{.}\hspace{.4pt}18653\discretionary{/}{%
}{/}v1\discretionary{/}{%
}{/}2020\hspace{.1pt}\discretionary{.}{%
}{.}\hspace{.4pt}emnlp\discretionary{%
}{-}{-}main\hspace{.1pt}\discretionary{.}{%
}{.}\hspace{.4pt}373}}}


\bibitem{singh2021com2sense}
S.~Singh, N.~Wen, Y.~Hou, P.~Alipoormolabashi, T.-L. Wu, X.~Ma, and N.~Peng.
\newblock Com2sense: A commonsense reasoning benchmark with complementary
  sentences.
\newblock In {\em Findings of ACL: ACL/IJCNLP}, pp. 883--898. ACL, online,
  2021. \href{https://doi.org/10.18653/v1/2021.findings-acl.78}
{doi: {{%
10\hspace{.1pt}\discretionary{.}{%
}{.}\hspace{.4pt}18653\discretionary{/}{%
}{/}v1\discretionary{/}{%
}{/}2021\hspace{.1pt}\discretionary{.}{%
}{.}\hspace{.4pt}findings\discretionary{%
}{-}{-}acl\hspace{.1pt}\discretionary{.}{%
}{.}\hspace{.4pt}78}}}


\bibitem{slack2021counterfactual}
D.~Slack, A.~Hilgard, H.~Lakkaraju, and S.~Singh.
\newblock Counterfactual explanations can be manipulated.
\newblock In {\em NeurIPS}, vol.~34, pp. 62--75, 2021.

\bibitem{conceptnet}
R.~Speer, J.~Chin, and C.~Havasi.
\newblock Conceptnet 5.5: An open multilingual graph of general knowledge.
\newblock In {\em AAAI}, pp. 4444--4451, 2017.

\bibitem{strobelt2017lstmvis}
H.~Strobelt, S.~Gehrmann, H.~Pfister, and A.~M. Rush.
\newblock Lstmvis: A tool for visual analysis of hidden state dynamics in
  recurrent neural networks.
\newblock {\em IEEE Trans. Visual Comput. Graphics}, 24(1):667--676, 2018.
  \href{https://doi.org/10.1109/TVCG.2017.2744158}
{doi: {{%
10\hspace{.1pt}\discretionary{.}{%
}{.}\hspace{.4pt}1109\discretionary{/}{%
}{/}TVCG\hspace{.1pt}\discretionary{.}{%
}{.}\hspace{.4pt}2017\hspace{.1pt}\discretionary{.}{%
}{.}\hspace{.4pt}2744158}}}


\bibitem{sundararajan2017axiomatic}
M.~Sundararajan, A.~Taly, and Q.~Yan.
\newblock Axiomatic attribution for deep networks.
\newblock In {\em Proc. ICML}, pp. 3319--3328. PMLR, 2017.

\bibitem{CSQA1}
A.~Talmor, J.~Herzig, N.~Lourie, and J.~Berant.
\newblock Commonsenseqa: A question answering challenge targeting commonsense
  knowledge.
\newblock In {\em Proc. NAACL}, pp. 4149--4158. ACL, Minneapolis, Minnesota,
  2019. \href{https://doi.org/10.18653/v1/n19-1421}
{doi: {{%
10\hspace{.1pt}\discretionary{.}{%
}{.}\hspace{.4pt}18653\discretionary{/}{%
}{/}v1\discretionary{/}{%
}{/}n19\discretionary{%
}{-}{-}1421}}}


\bibitem{CSQA2}
A.~Talmor, O.~Yoran, R.~L. Bras, C.~Bhagavatula, Y.~Goldberg, Y.~Choi, and
  J.~Berant.
\newblock Commonsenseqa 2.0: Exposing the limits of {AI} through gamification.
\newblock In {\em NeurIPS Datasets and Benchmarks Track}, vol.~1, 2021.

\bibitem{webchild}
N.~Tandon, G.~de~Melo, F.~Suchanek, and G.~Weikum.
\newblock Webchild: Harvesting and organizing commonsense knowledge from the
  web.
\newblock In {\em Proc. WSDM}, p. 523–532. ACM, New York, 2014.
  \href{https://doi.org/10.1145/2556195.2556245}
{doi: {{%
10\hspace{.1pt}\discretionary{.}{%
}{.}\hspace{.4pt}1145\discretionary{/}{%
}{/}2556195\hspace{.1pt}\discretionary{.}{%
}{.}\hspace{.4pt}2556245}}}


\bibitem{tenney2019bert}
I.~Tenney, D.~Das, and E.~Pavlick.
\newblock Bert rediscovers the classical nlp pipeline.
\newblock In {\em Proc. ACL}, pp. 4593--4601. ACL, Florence, Italy, 2019.
  \href{https://doi.org/10.18653/v1/p19-1452}
{doi: {{%
10\hspace{.1pt}\discretionary{.}{%
}{.}\hspace{.4pt}18653\discretionary{/}{%
}{/}v1\discretionary{/}{%
}{/}p19\discretionary{%
}{-}{-}1452}}}


\bibitem{tenney2020language}
I.~Tenney, J.~Wexler, J.~Bastings, T.~Bolukbasi, A.~Coenen, S.~Gehrmann,
  E.~Jiang, M.~Pushkarna, C.~Radebaugh, E.~Reif, et~al.
\newblock The language interpretability tool: Extensible, interactive
  visualizations and analysis for nlp models.
\newblock In {\em Proc. EMNLP: System Demonstrations}, pp. 107--118, 2020.
  \href{https://doi.org/10.18653/v1/2020.emnlp-demos.15}
{doi: {{%
10\hspace{.1pt}\discretionary{.}{%
}{.}\hspace{.4pt}18653\discretionary{/}{%
}{/}v1\discretionary{/}{%
}{/}2020\hspace{.1pt}\discretionary{.}{%
}{.}\hspace{.4pt}emnlp\discretionary{%
}{-}{-}demos\hspace{.1pt}\discretionary{.}{%
}{.}\hspace{.4pt}15}}}


\bibitem{vig2019bertviz}
J.~Vig.
\newblock Bertviz: A tool for visualizing multihead self-attention in the bert
  model.
\newblock In {\em ICLR Workshop: Debugging Machine Learning Models}, 2019.

\bibitem{voita-titov-2020-information}
E.~Voita and I.~Titov.
\newblock Information-theoretic probing with minimum description length.
\newblock In {\em Proc. EMNLP}, pp. 183--196. ACL, Online, Nov. 2020.
  \href{https://doi.org/10.18653/v1/2020.emnlp-main.14}
{doi: {{%
10\hspace{.1pt}\discretionary{.}{%
}{.}\hspace{.4pt}18653\discretionary{/}{%
}{/}v1\discretionary{/}{%
}{/}2020\hspace{.1pt}\discretionary{.}{%
}{.}\hspace{.4pt}emnlp\discretionary{%
}{-}{-}main\hspace{.1pt}\discretionary{.}{%
}{.}\hspace{.4pt}14}}}


\bibitem{wang2021m2lens}
X.~Wang, J.~He, Z.~Jin, M.~Yang, Y.~Wang, and H.~Qu.
\newblock M2lens: Visualizing and explaining multimodal models for sentiment
  analysis.
\newblock {\em IEEE Trans. Visual Comput. Graphics}, 28(1):802--812, 2021.
  \href{https://doi.org/10.1109/TVCG.2021.3114794}
{doi: {{%
10\hspace{.1pt}\discretionary{.}{%
}{.}\hspace{.4pt}1109\discretionary{/}{%
}{/}TVCG\hspace{.1pt}\discretionary{.}{%
}{.}\hspace{.4pt}2021\hspace{.1pt}\discretionary{.}{%
}{.}\hspace{.4pt}3114794}}}


\bibitem{DBLP:conf/emnlp/WeiZ19}
J.~W. Wei and K.~Zou.
\newblock {EDA:} easy data augmentation techniques for boosting performance on
  text classification tasks.
\newblock In {\em Proc. EMNLP}, pp. 6381--6387. ACL, Hong Kong, 2019.
  \href{https://doi.org/10.18653/v1/D19-1670}
{doi: {{%
10\hspace{.1pt}\discretionary{.}{%
}{.}\hspace{.4pt}18653\discretionary{/}{%
}{/}v1\discretionary{/}{%
}{/}D19\discretionary{%
}{-}{-}1670}}}


\bibitem{whatiftool}
J.~Wexler, M.~Pushkarna, T.~Bolukbasi, M.~Wattenberg, F.~Vi{\'e}gas, and
  J.~Wilson.
\newblock The what-if tool: Interactive probing of machine learning models.
\newblock {\em IEEE Trans. Visual Comput. Graphics}, 26(1):56--65, 2019.
  \href{https://doi.org/10.1109/TVCG.2019.2934619}
{doi: {{%
10\hspace{.1pt}\discretionary{.}{%
}{.}\hspace{.4pt}1109\discretionary{/}{%
}{/}TVCG\hspace{.1pt}\discretionary{.}{%
}{.}\hspace{.4pt}2019\hspace{.1pt}\discretionary{.}{%
}{.}\hspace{.4pt}2934619}}}


\bibitem{wu2021polyjuice}
T.~Wu, M.~T. Ribeiro, J.~Heer, and D.~S. Weld.
\newblock Polyjuice: Generating counterfactuals for explaining, evaluating, and
  improving models.
\newblock In {\em Proc. ACL}, pp. 6707--6723. ACL, Online, 2021.
  \href{https://doi.org/10.18653/v1/2021.acl-long.523}
{doi: {{%
10\hspace{.1pt}\discretionary{.}{%
}{.}\hspace{.4pt}18653\discretionary{/}{%
}{/}v1\discretionary{/}{%
}{/}2021\hspace{.1pt}\discretionary{.}{%
}{.}\hspace{.4pt}acl\discretionary{%
}{-}{-}long\hspace{.1pt}\discretionary{.}{%
}{.}\hspace{.4pt}523}}}


\bibitem{humanparity}
Y.~Xu, C.~Zhu, S.~Wang, S.~Sun, H.~Cheng, X.~Liu, J.~Gao, P.~He, M.~Zeng, and
  X.~Huang.
\newblock Human parity on commonsenseqa: Augmenting self-attention with
  external attention.
\newblock In {\em Proc. IJCAI}, pp. 2762--2768, 2022.
  \href{https://doi.org/10.24963/ijcai.2022/383}
{doi: {{%
10\hspace{.1pt}\discretionary{.}{%
}{.}\hspace{.4pt}24963\discretionary{/}{%
}{/}ijcai\hspace{.1pt}\discretionary{.}{%
}{.}\hspace{.4pt}2022\discretionary{/}{%
}{/}383}}}


\bibitem{yasunaga2021qagnn}
M.~Yasunaga, H.~Ren, A.~Bosselut, P.~Liang, and J.~Leskovec.
\newblock Qa-gnn: Reasoning with language models and knowledge graphs for
  question answering.
\newblock In {\em Proc. NAACL}, pp. 535--546. ACL, online, 2021.
  \href{https://doi.org/10.18653/v1/2021.naacl-main.45}
{doi: {{%
10\hspace{.1pt}\discretionary{.}{%
}{.}\hspace{.4pt}18653\discretionary{/}{%
}{/}v1\discretionary{/}{%
}{/}2021\hspace{.1pt}\discretionary{.}{%
}{.}\hspace{.4pt}naacl\discretionary{%
}{-}{-}main\hspace{.1pt}\discretionary{.}{%
}{.}\hspace{.4pt}45}}}


\bibitem{zellers2019hellaswag}
R.~Zellers, A.~Holtzman, Y.~Bisk, A.~Farhadi, and Y.~Choi.
\newblock Hellaswag: Can a machine really finish your sentence?
\newblock In {\em Proc. ACL}, pp. 4791--4800. ACL, Florence, Italy, 2019.
  \href{https://doi.org/10.18653/v1/p19-1472}
{doi: {{%
10\hspace{.1pt}\discretionary{.}{%
}{.}\hspace{.4pt}18653\discretionary{/}{%
}{/}v1\discretionary{/}{%
}{/}p19\discretionary{%
}{-}{-}1472}}}


\bibitem{transomcs}
H.~Zhang, D.~Khashabi, Y.~Song, and D.~Roth.
\newblock Transomcs: From linguistic graphs to commonsense knowledge.
\newblock In {\em Proc. IJCAI}, pp. 4004--4010, 2020.
  \href{https://doi.org/10.24963/ijcai.2020/554}
{doi: {{%
10\hspace{.1pt}\discretionary{.}{%
}{.}\hspace{.4pt}24963\discretionary{/}{%
}{/}ijcai\hspace{.1pt}\discretionary{.}{%
}{.}\hspace{.4pt}2020\discretionary{/}{%
}{/}554}}}


\bibitem{zhou2020evaluating}
X.~Zhou, Y.~Zhang, L.~Cui, and D.~Huang.
\newblock Evaluating commonsense in pre-trained language models.
\newblock In {\em AAAI}, vol.~34, pp. 9733--9740, 2020.
  \href{https://doi.org/10.1609/aaai.v34i05.6523}
{doi: {{%
10\hspace{.1pt}\discretionary{.}{%
}{.}\hspace{.4pt}1609\discretionary{/}{%
}{/}aaai\hspace{.1pt}\discretionary{.}{%
}{.}\hspace{.4pt}v34i05\hspace{.1pt}\discretionary{.}{%
}{.}\hspace{.4pt}6523}}}


\end{thebibliography}


\newpage

\appendix 
\section{Model editing implementation}\label{sec.model_editing}
Here, we introduce the details of technical implementation and evaluation of editor networks for model editing. We adopt the MEND network~\cite{mitchell2022fast} for model editing.


\subsection{Editor networks}

MEND leverages a collection of small auxiliary editing networks that use a single desired input-output pair to make fast, local edits to pretrained models. 
Specifically, it uses the low-rank structure of fine-tuning gradients to enable scalable and efficient editing of very large pretrained language models on specified layers of transformers.
MEND uses the fact that gradients for MLPs are rank-1 matrix and apply the theory to Transformers by summing elements over sequence indices. 
The model editing gradient update function is derived as:

$$\tilde{\nabla}_{W_\ell} = \sum_{i=1}^B\tilde\delta_{\ell + 1}^{i}{\tilde u_\ell^{i\,\top}}.$$

Where $\tilde u_\ell^i$ and $\tilde\delta_{\ell+1}^i$ are \textit{pseudo-activations}  and \textit{pseudo-delta} by taking the sequence sum of the gradient of the loss for batch $i$ with respect to the pre-activations at layer $l$ + 1, and the sequence sum of the inputs to layer $l$ for batch element $i$. $B$ is the number of total batches.
$\tilde{\nabla}_{W_\ell}$ is the gradient update to be applied on the MLP layers of transformers.
For more details, please refer to the original paper~\cite{mitchell2022fast}.


\subsection{Model editing training}

Practically, for the T5-based QA model that we use \footnote{\url{https://github.com/allenai/unifiedqa}}, we only edit the MLP layers of the last two encoder and decoder blocks of the transformer.
We follow the official implementation of MEND\footnote{\url{https://github.com/eric-mitchell/mend/}} to build our model and conduct the experiments.

To train editor networks that can edit our T5 model on CSQA, we need to collect editing targets, equivalence examples, and locality examples. Specifically, editing targets contain a question and a target choice, where the question comes from train set of CSQA dataset, and the target choice does not necessarily be the ground truths. We randomly sample one choice from five alternatives in the original QA instance as the editing target.
For equivalence examples generation, we use data augmentation techniques to perturbate the original instances to get meaning-preserving augmentations as much as possible.
We use back-translation implemented in \texttt{nlpaug}\footnote{\url{https://github.com/makcedward/nlpaug}} to translate the original sentences to German and then back to English using the \texttt{facebook/wmt19-en-de} machine translation checkpoint. We also adopt Easy Data Augmentation (EDA)~\cite{DBLP:conf/emnlp/WeiZ19} to do synonym replacement and random insert/delete/replace on the original sentences to ensure the robustness of the model training.
For locality examples, we independently sample negative examples different from editing targets from the same original dataset.

Once the editor networks are trained, they can be applied to conduct posthoc editing on the original model at inference time.

\section{System interaction designs}
\label{sys.interactions}
\name{} offers various interactions to support multi-level analysis of model behavior with details on demand.


\textbf{Lasso and pan-and-zoom}. 
In the \gv{}, users can lasso a group of data instances in the scatter plots to examine the details in the \sv{} and \iv{}. 
And users can use pan-and-zoom in the scatter plots to navigate local clusters more easily. 


\textbf{Hovering and clicking}. 
To make the interface cleaner and less overwhelming, we hide lots of details, and users can hover or click to see the details on demand. For example, in the \gv{}, users can hover over
the dots in the scatter plots and the relation bars in the middle to see the pairs of question stems and target concepts and relation accuracy, respectively.
When hovering the cluster glyphs in the \sv{}, detailed concepts and statistics of the clusters, together with their connections with other clusters, will be displayed. 
In the \iv{}, hovering over the charts will display the detailed numbers. Also, users can hover over the answer choices to query their relations with the question stem concepts.

Moreover, users can filter or highlight the information by clicking.
For example, relation bars in the \gv{}, and stacked bars in the \sv{} can be clicked to filter data instances.
In addition, users can navigate through instances by clicking the pagination buttons. Meanwhile, its corresponding dots and clusters will be highlighted in the \gv{} and \sv, respectively.




\section{Evaluation of Commonsense Knowledge Coverage of \cpn{}}
We conducted an evaluation using a random sample of 100 examples drawn from the CSQA validation set\footnote{Data samples fo evaluation are available at \url{https://bit.ly/3PCGbze}}.
We have invited an NLP expert (\imp{E14}, not our co-author) to evaluate the relational paths extracted by our algorithm based on ConceptNet. 
For each QA instance, the expert examined the QA instance and the extracted relational paths built by retrieved concept-relation triplets from ConceptNet.
Then, he decided whether the paths could accurately cover the necessary commonsense knowledge to answer the question. 
Finally, we calculated and reported the proportion of instances for which the necessary commonsense knowledge is covered by the extracted ConceptNet knowledge. 

The results show that the retrieved ConceptNet knowledge can cover the commonsense knowledge in 91 out of 100 instances. 
It further helps validate the use of \cpn{} for model contextualization on the CSQA dataset. 
However, although CSQA is built upon \cpn{}, it still cannot cover some commonsense knowledge in the data.
For example, for the question \hl{The potato might be the official vegetable of what? (correct answer: Maryland)}, retrieved concept-relation triplets from \cpn{} fail to build a connection between ``potato'' and ``Maryland'' or ``official vegetable''.
In addition, for the question \hl{Where has the newest baseball stadium? (correct answer: Phoenix)}, although retrieved concept-relation triplets can associate ``baseball stadium`` with different locations using \texttt{AtLocation} realation. However, it lacks the knowledge to determine which city has the ``newest'' stadium.

\section{Additional Cases of Using \name{}}
\label{app.case_studies}
\subsection{Relation of \texttt{atlocation} regarding room and office is relatively well-learned}\label{app.case1}

\begin{figure*}[ht]
    \includegraphics[width=\textwidth]{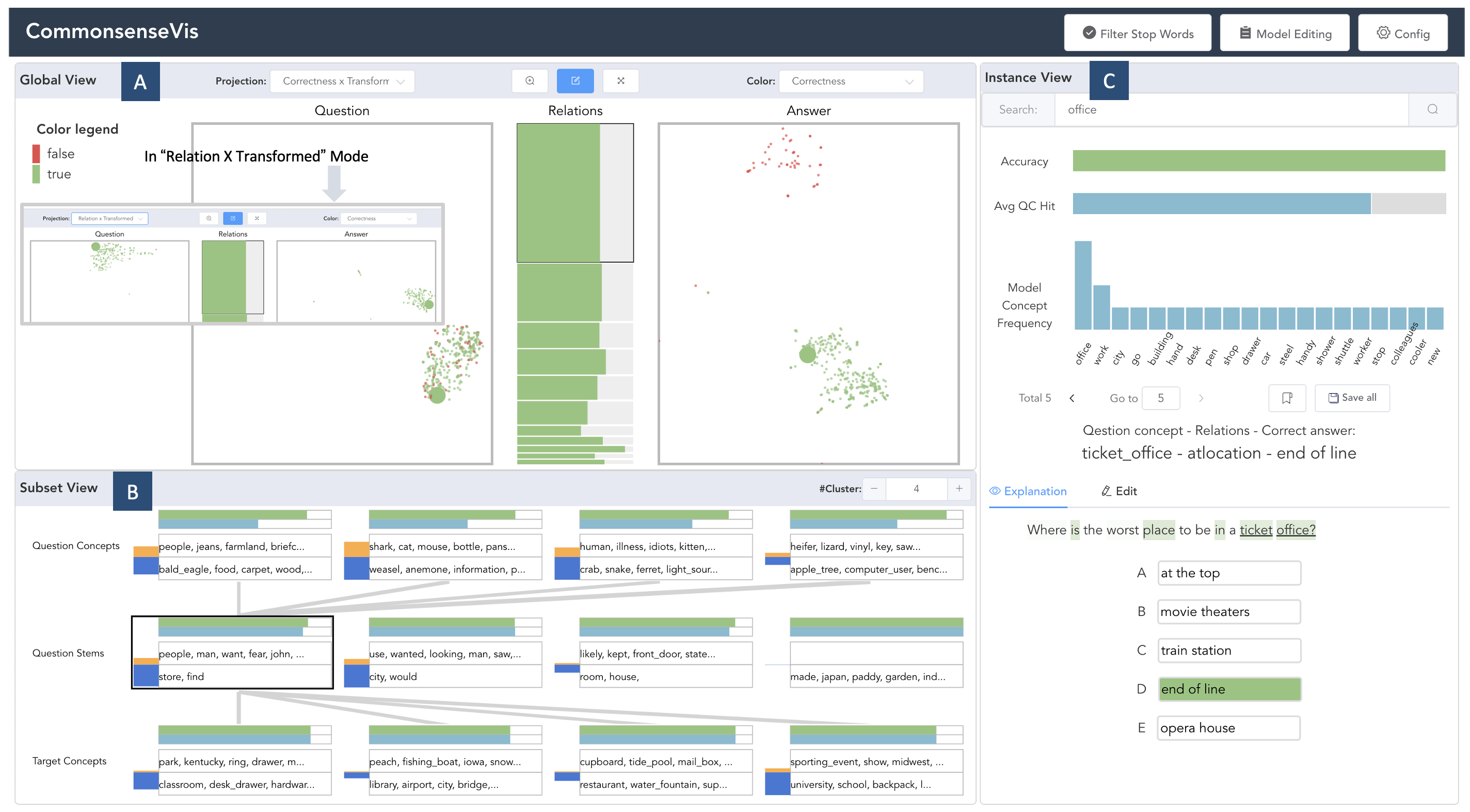}
    \caption{\renfei{Interactive exploration of case three by \imp{E4} using \name{}. (A) \imp{E4} used ``\textit{Correctness X Tranformed}'' projection mode to select the correct instances and changed into ``\textit{Relation X Tranformed}'' mode to visualize the corresponding result. (B) \imp{E4} went through the cluster glyphs in the \sv{} to gain general information about the model's behavior over correctly answered subsets. (C) After clicking on a cluster, \imp{E4} was able to check more statistics about this group on the top (the accuracy, the average question concept (QC) coverage ratio, and the frequent model concepts). Then \imp{E4} quickly went through pages to check the detailed model behavior, such as feature importance scores, to check what kind of commonsense knowledge the model learns sufficiently. }}
    \label{fig:case1 full}
\end{figure*}

\textbf{Global Summary}
(\imp{R1, R2}) 
After loading the system and dataset, the expert \imp{E4} first referred to the \gv{} to explore the model performance regarding different relations. 
After hovering over the green bars between the scatter plots, he was able to quickly observe that although there is an imbalanced relation distribution (as indicated by varied bar height), accuracies for most relations are about 0.70 (\renfei{\autoref{fig:case1 full}A}). It indicates that the model may have learned a fair amount of relations between different concepts.
Then, \imp{E4} felt curious about what relations are and under what contexts the model learns well. 
He started with the relation \texttt{atlocation} with the highest green bar at the top. After clicking the bar, he selected the ``\textit{Correctness X Tranformed}'' projection mode and ``\textit{Correctness}'' coloring scheme to explore the distribution of the correctly-answered instances of \texttt{atlocation} in the question and answer (target concept) scatter plots (\renfei{\autoref{fig:case1 full}A}). 
He noticed that there is a large cluster with green dots in the answer scatter plot. He wondered whether the models have really learned \texttt{atlocation} between question and target concepts in these instances. Therefore, he switched to ``\textit{Relation X Tranformed}'' projection mode to see how the relation is learned by examining the correspondence between question stems and target concepts after transformation (\renfei{\autoref{fig:case1 full}A}). And he discovered two well-formed and well-aligned clusters in the two scatter plots, which provides support for a good learning of \texttt{atlocation} relation.

\textbf{Subset Exploration}
(\imp{R1, R2, R3}) 
To further explore the contexts of selected instances, \imp{E4} looked at the question stem cluster glyphs in the \sv{} (\renfei{\autoref{fig:case1 full}B}), where \xingbo{the green and blue bars} nearly occupy the two stacked bars at the top. It indicates a high model accuracy and overlap between the model concepts and ConceptNet concepts.
Moreover, he observed the yellow rectangles on the left of question stem clusters are much shorter than the dark blue ones (\renfei{\autoref{fig:case1 full}B}), confirming that very few ConceptNet concepts are not covered by the model. 
He then hovered over the first cluster glyph to see the details of those concepts, where words like ``man'' and ``want'' appear. He thought that these concepts, not important to model predictions, might not affect the reasoning about \texttt{atlocation}.
Therefore, he hypothesized that the question contexts are properly considered by the model.
And he clicked this glyph to explore detailed instances and their explanations in the \iv{} to verify his hypothesis.
By scanning the top frequent model concepts in the histogram (\renfei{\autoref{fig:case1 full}C}) (\eg, ``where'', ``what'', ``store'', ``office'', ``room'', ``building''), he reasoned that many of these instances of \texttt{atlocation} are ``what'', ``where'' questions and associate with ``office'' and ``room''.

\begin{figure}[h]
\centering
    \includegraphics[width=\linewidth]{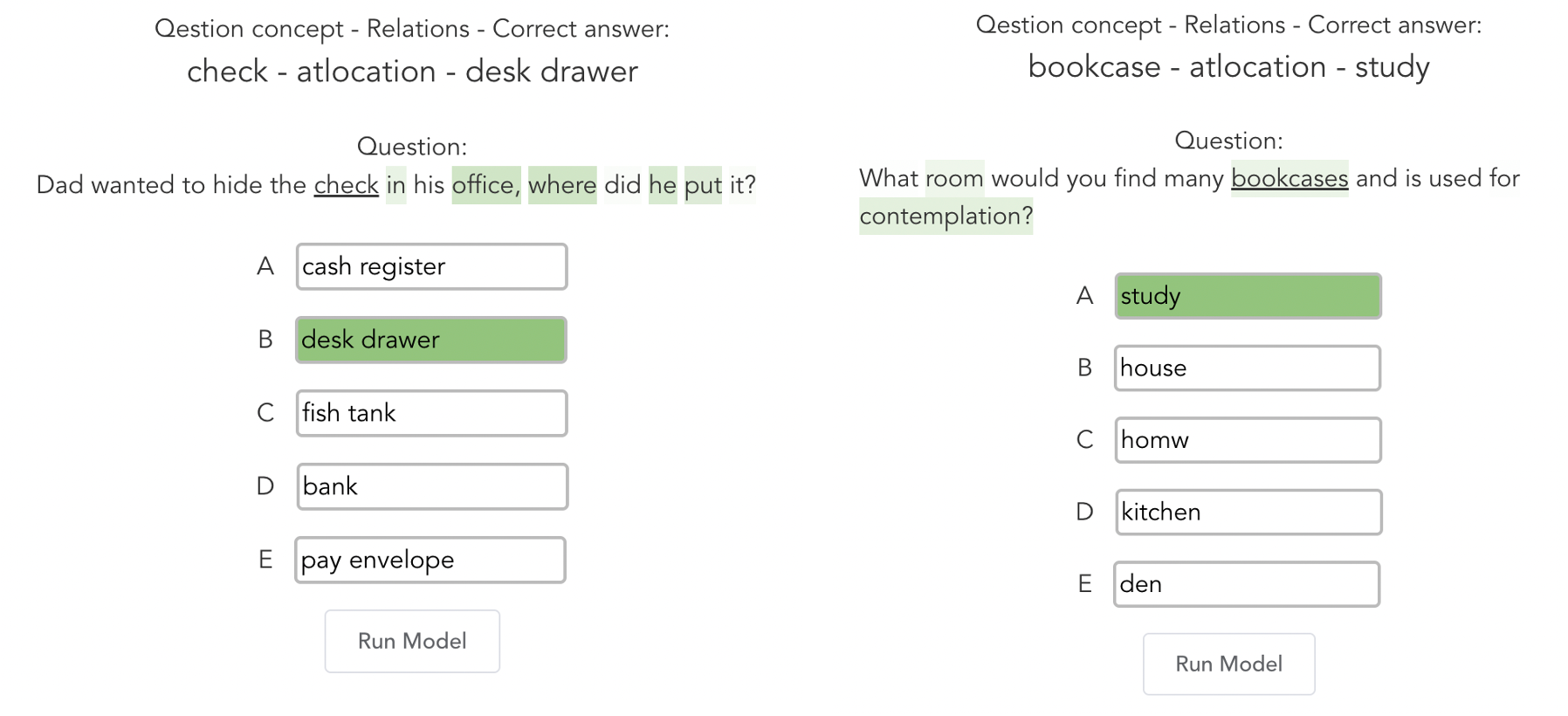}
    \caption{\renfei{Instances found by \imp{E4} confirmed his hypothesis that the relation of \texttt{atlocation} regarding scenarios about room and office is well-learned by the model. }}
    \label{fig:case1 instances}
\end{figure}

\textbf{Instance Exploration and Searching}
(\imp{R1, R4}) 
Finally, through exploration of individual questions in the \iv{}, \imp{E4} found that the model truly captures important words for answering commonsense questions.
For example, in \renfei{\autoref{fig:case1 instances}}, SHAP values show ``office'' and ``put'' as important contexts for where the ``check'' can be located, which is ``desk drawer''. 
Another example in \renfei{\autoref{fig:case1 instances}} shows that the model properly considered contexts like ``room'' and ``contemplation'' for where the ``bookcases'' should be located in a ``study room'', which aligns with human knowledge. 
Then, \imp{E4} reasoned that the model has a good sense of \texttt{atlocation} in the situations of ``office'' and ``room''. 
And he lassoed all the instances of \texttt{atlocation} in the \gv{} and typed ``office'' and ``room'' to search relevant instances in the \iv{}, where the model achieves 90.00\% and 88.89\% accuracy, respectively (much higher than the overall 71.00\% accuracy). 
\imp{E4} was convinced that the model has learned the relation \texttt{atlocation} in the context of ``office'' and ``room'' properly.



\begin{figure}[h]
\vspace{-5mm}
    \includegraphics[width=\columnwidth]{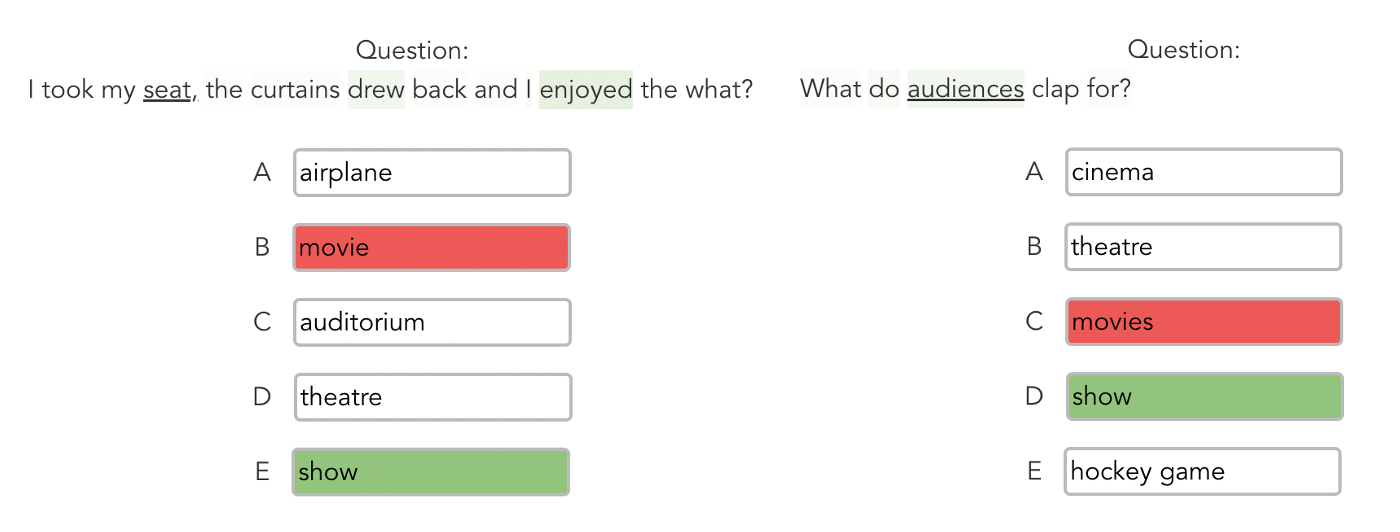}
    \caption{\renfeivis{Case two: Instances related to ``movie'' found by \imp{E5}. }}
    \label{fig:case3 movie}
\end{figure}

\section{Additional User Study Results}
\label{app.user_study}

\subsection{User study questionnaire}
\label{app.user_study_question}
The user study questionnaire is presented in \autoref{tab:questionnaire-table}.
\begin{table}[!htb]
\caption{The first section of our questionnaire is designed to collect feedback on the system's effectiveness in evaluating the model's commonsense abilities (Q1-Q4). The second section is designed to evaluate the usefulness and usability of \gv{} (Q5-Q7), \sv{} (Q8-Q10) and \iv{} (Q11-Q13). The last section is designed to evaluate personal opinions of our system (Q14-Q17). The original sentences without the words in brackets are the positive statements at the right end of the scale points, while the sentences with words in the brackets are the negative statements at the left end. }
\label{tab:questionnaire-table}
\resizebox{0.9\linewidth}{!}{
\begin{tabular}{p{0.5cm}|p{0.85\linewidth}}

\toprule
Q1 & The system can (cannot) help me identify the target commonsense knowledge in data instances. \\ 
Q2 & The system does (does not) contextualize model performance regarding different concepts and their underlying relations. \\ 
Q3 & The system can (cannot) help me infer the model’s relational reasoning over different concepts. \\ 
Q4 & I am (not) confident in my findings about the model’s commonsense reasoning abilities. \\ \hline
Q5 & The Global View can (cannot) help me relate model performance to different concepts and relations. \\ 
Q6 & The Global View can (cannot) help me infer how the relations are learned by models. \\ 
Q7 & The Global View is easy (difficult) to understand. \\ \hline
Q8 & The Subset View can (cannot) help me align model behavior with ConceptNet knowledge. \\ 
Q9 & The Subset View can (cannot) help me summarize model behavior on different groups of question concepts/question stems/target concepts. \\ 
Q10 & The Subset View is easy (difficult) to understand. \\ \hline
Q11 & The Instance View can (cannot) help me diagnose if the model uses proper information for reasoning. \\
Q12 & The Instance View can (cannot) help me infer if a relation between question concepts and target concepts is learned or not. \\
Q13 & The Instance View is easy (difficult) to understand. \\ \hline
Q14 & It is easy (difficult) to learn the system. \\
Q15 & It is easy (difficult) to use the system. \\
Q16 & I will (will not) use it in the future for understanding and diagnosing language models. \\
Q17 & I will (will not) recommend this system to other colleagues for understanding and diagnosing language models. \\ \bottomrule
\end{tabular}
}
\end{table}

\subsection{User ratings and feedback}
\label{app.user_rating_feedback}
\subsubsection{Visual designs and interactions}
\begin{figure}[t]
\centering
    \includegraphics[width=.8\linewidth]{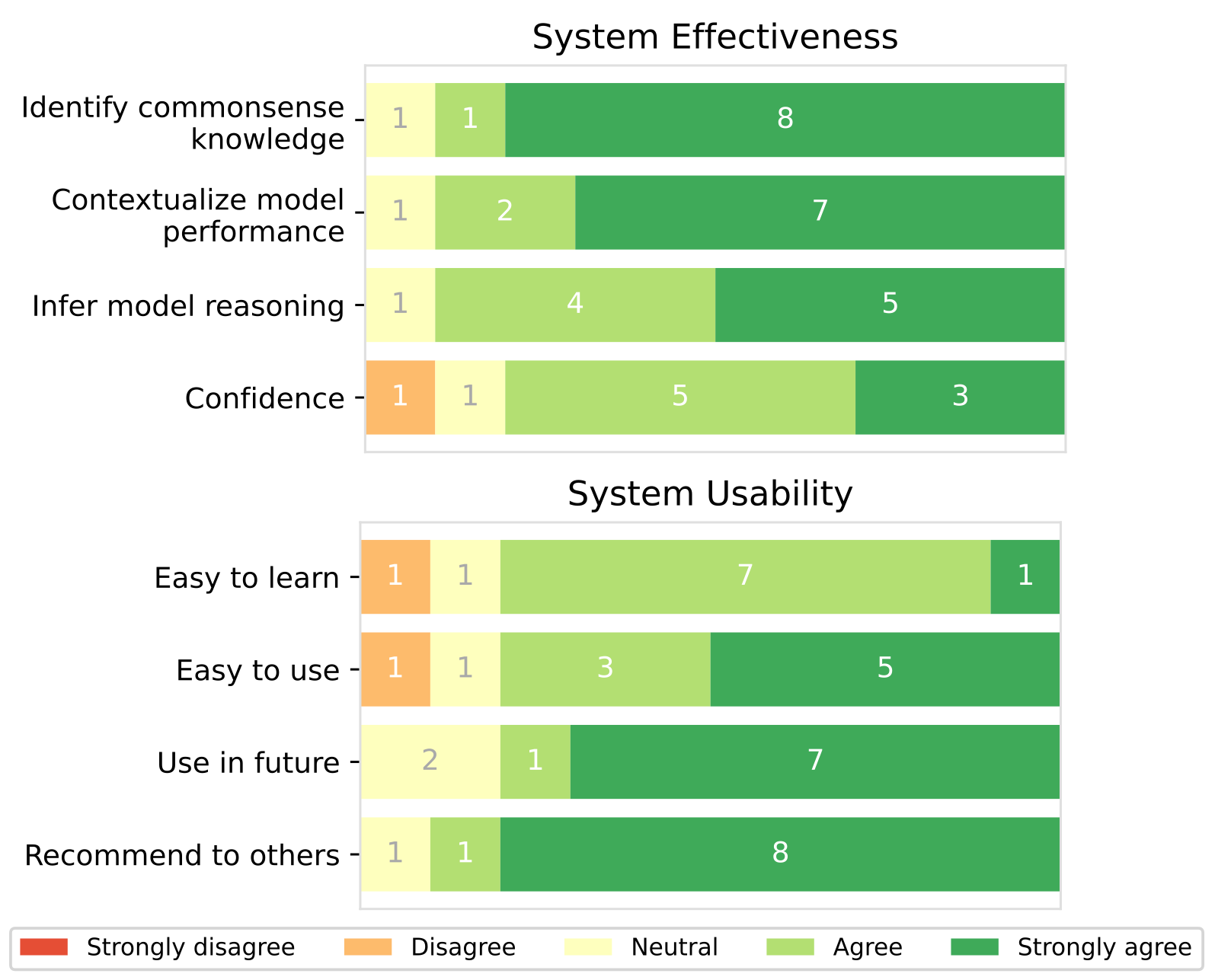}
    \caption{The results of the questionnaire about overall impressions of our system, including the effectiveness (Q1-Q4) and the usability (Q14-Q17). }
    \label{fig:system-user-scores}
\end{figure}

\begin{figure}[h]
\centering
    \includegraphics[width=.8\linewidth]{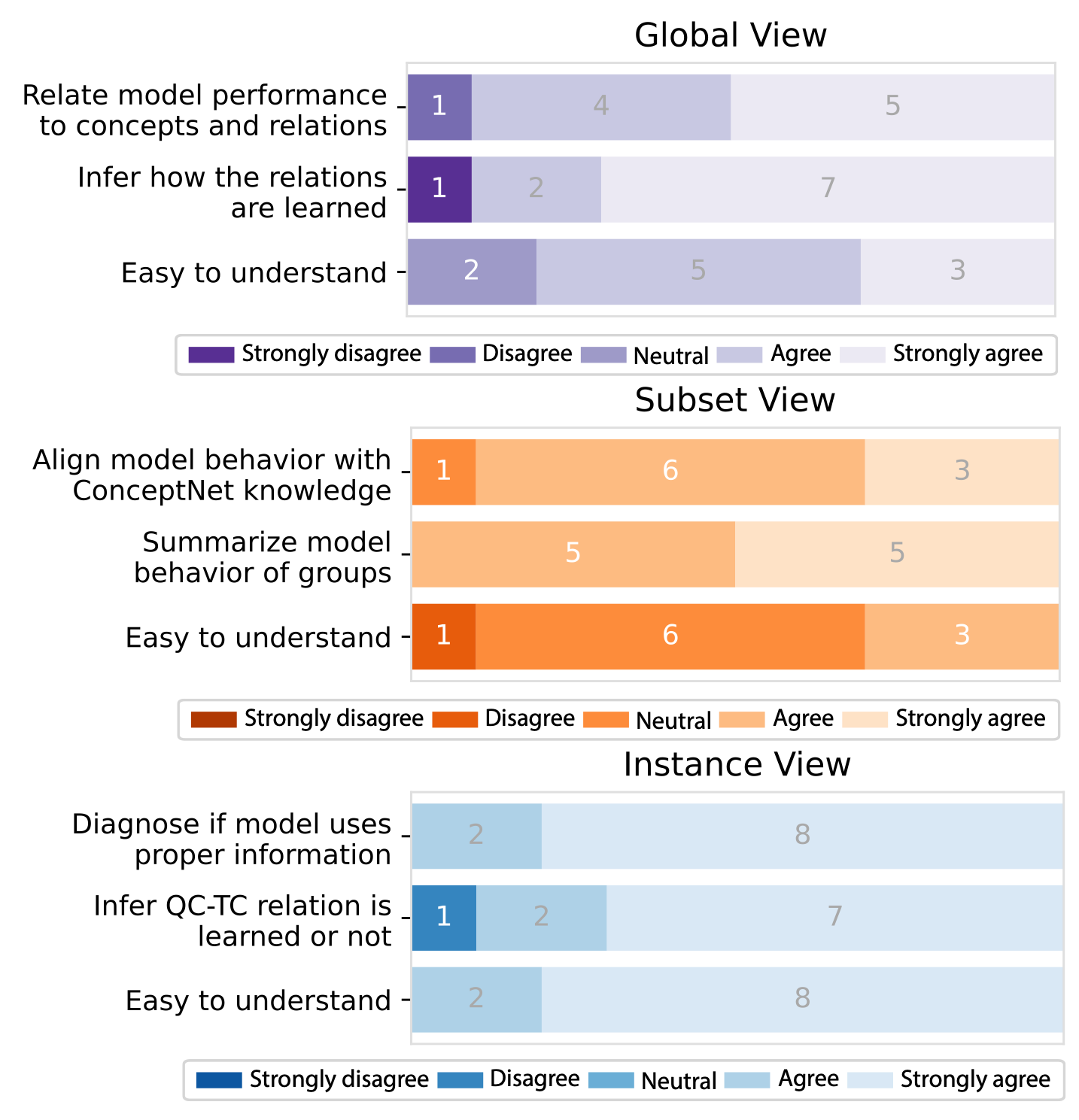}
    \caption{The results of the questionnaire about the helpfulness and intuitiveness of using our system to evaluate commonsense reasoning abilities regarding the \gv{} (Q5-Q7), the \sv{} (Q8-Q10), and the \iv{} (Q11-Q13). }
    \label{fig:three-view-user-scores}
\end{figure}

As shown in \autoref{fig:system-user-scores},  participants generally agreed that our system is easy to use ($Mean_{Q15} = 4.20$, $SD_{Q15} = 1.03$) while it required some efforts for learning ($Mean_{Q14} = 3.80$, $SD_{Q14} = 0.79$).
They were willing to use ($Mean_{Q16} = 4.50$, $SD_{Q16} = 0.85$) and recommend our system ($Mean_{Q17} = 4.70$, $SD_{Q17} = 0.67$) for understanding and diagnosing commonsense reasoning capabilities of language models.
The most intuitive view of \name{} is the \iv{}, then the \gv{}. And the \sv{} was thought to be the most difficult to understand. 
We summarize participants' feedback (as shown in \autoref{fig:three-view-user-scores}) on our visual designs as follows.

For the \gv{}, participants found it quite useful for finding relations/concepts with large/small prediction errors ($Mean_{Q5} = 4.30$, $SD_{Q5} = 0.95$). \hl{I can quickly observe the correctness distribution among instances} (\imp{E6, E8, E12}) and \hl{narrow down to specific cluster of instances} (\imp{E7, E9}). 
And the question and answer scatter plots helped them infer if the relations are generally learned well ($Mean_{Q6} = 4.40$, $SD_{Q6} = 1.26$). 
Furthermore, \imp{E6} and \imp{E8} added that the correctness coloring of the dots (\ie, model accuracy) is really helpful when they analyze relation learning and were not sure about the quality of the alignment between questions and answers. 
However, some participants reported that sometimes it is a bit hard to visually align and match clusters of dots in the question and answer scatter plots due to the embedding rotation effect (\imp{E10}) and scarcity of instances (\imp{E11}).

For the \sv{}, participants thought it was a bit complex ($Mean_{10} = 3.20$, $SD_{Q10} = 0.63$). And after they got familiar with it, they considered it helpful to compare the model concepts and ConceptNet concepts ($Mean_{Q8} = 4.20$, $SD_{Q8} = 0.63$) and summarize model behavior on different question concepts/question stems/target concepts ($Mean_{Q9} = 4.50$, $SD_{Q9} = 0.53$).
For example, \imp{E9} commented \hl{The hit ratio and top missed concepts are very useful in understanding what types of concepts the model focuses on.}
Some participants felt \sv{} could be a bit too informative sometimes. 
\imp{E10} suggested, \hl{Choosing the number of clusters and viewing the concepts and linkers are a bit messy and too informative, can we simplify the design and only the most impactful one?}

The \iv{} was the most favored by participants for its intuitiveness and helpfulness in diagnosing if the model uses proper information ($Mean_{Q11} = 4.80$, $SD_{Q11} = 0.42$) and learns relations between question concepts and target concepts for reasoning ($Mean_{Q12} = 4.50$, $SD_{Q12} = 0.97$). 
Participants generally thought that SHAP explanations and model probing complement each other to deepen the model understanding.
\hl{I tested many examples and found that the explanation results were very satisfactory. And the instance probing also helped me to do some further investigation and testing on model behavior.} (\imp{E7}).
\hl{The model probing is a great tool to change the input to the model and check the behavioral change of the model. This can be used to do some causal analysis of concepts.} (\imp{E8}).






\end{document}